\documentclass[10pt,journal,compsoc,twoside]{IEEEtran}

\usepackage{cite}
\usepackage{booktabs}
\usepackage{multirow}
\usepackage{graphicx}
\usepackage{amssymb}
\usepackage{amsmath}
\usepackage{dsfont}
\usepackage{dsfont}
\usepackage{graphicx}
\usepackage{xcolor}
\usepackage{color,soul}
\usepackage{mathrsfs}
\usepackage{amsmath}  
\usepackage{colortbl}
\usepackage{enumitem}
\usepackage{arydshln}

\begin{document}



\makeatletter
\def\ps@IEEEtitlepagestyle{
  \def\@oddfoot{\mycopyrightnotice}
  \def\@evenfoot{\mycopyrightnotice}
}
\def\mycopyrightnotice{
  {\@IEEEheaderstyle
  \begin{minipage}{\textwidth}
  \centering
  \vspace{0.5cm}
  \copyright~2019 IEEE. Personal use of this material is permitted. Permission from IEEE must be obtained for all other uses, in any current or future media, including reprinting / republishing this material for advertising or promotional purposes, creating new collective works, for resale or redistribution to servers or lists, or reuse of any copyrighted component of this work in other works.
  \end{minipage}
  }
}

\title{Jointly Aligning and Predicting Continuous Emotion Annotations}

\author{
  Soheil~Khorram,
  Melvin~G~McInnis,
  Emily~Mower~Provost
  \IEEEcompsocitemizethanks{
    \IEEEcompsocthanksitem 
    Soheil~Khorram is a Research Fellow in the Department of Computer Science and Engineering, and the Department of Psychiatry, University of Michigan.
    E-mail: \texttt{khorram.soheil@gmail.com}
    \IEEEcompsocthanksitem
    Melvin~G~McInnis is the Thomas B and Nancy Upjohn Woodworth Professor of Bipolar Disorder and Depression, Department of Psychiatry, University of Michigan School of Medicine.
    \protect\\
    E-mail: \texttt{mmcinnis@umich.edu}
    \IEEEcompsocthanksitem
    Emily~Mower~Provost is an Associate Professor in the Department of Computer Science and Engineering, University of Michigan.
    \protect\\
    E-mail: \texttt{emilykmp@umich.edu}
  }
}

%
%

\markboth{IEEE Transactions on Affective Computing}{Khorram \MakeLowercase{\textit{et al.}}: Jointly Aligning and Predicting Continuous Emotion Annotations}

\IEEEtitleabstractindextext{%
\begin{abstract}
Time-continuous dimensional descriptions of emotions (e.g., arousal, valence) allow researchers to characterize short-time changes and to capture long-term trends in emotion expression. However, continuous emotion labels are generally not synchronized with the input speech signal due to delays caused by reaction-time, which is inherent in human evaluations. To deal with this challenge, we introduce a new convolutional neural network (\emph{multi-delay sinc network}) that is able to simultaneously align and predict labels in an end-to-end manner. The proposed network is a stack of convolutional layers followed by an aligner network that aligns the speech signal and emotion labels. This network is implemented using a new convolutional layer that we introduce, the \emph{delayed sinc layer}. It is a time-shifted low-pass (sinc) filter that uses a gradient-based algorithm to learn a single delay. Multiple delayed sinc layers can be used to compensate for a non-stationary delay that is a function of the acoustic space. 
We test the efficacy of this system on two common emotion datasets, RECOLA and SEWA, and show that this approach obtains state-of-the-art speech-only results by learning time-varying delays while predicting dimensional descriptors of emotions. 

\end{abstract}

}
\maketitle
\setcounter{page}{1}
\IEEEdisplaynontitleabstractindextext
\IEEEpeerreviewmaketitle

%

\IEEEraisesectionheading{\section{Introduction}\label{sec:introduction}}

\IEEEPARstart{E}{motions} are complex and dynamic manifestations of internal human experience.  Descriptions of emotion have often relied upon categories  (e.g. happiness, anger, or disgust~\cite{ekman1992argument, gideon2016wild}).  However, these categories have substantial limitations due to the influence of cultural and other types of context~\cite{cordaro2018universals}. As a result, the field of affective computing has increasingly adopted dimensional measures to quantify emotional expressions~\cite{schuller2018speech, mencattini2017continuous, ringeval2015av+, aldeneh2017pooling, keren2017end, abdelwahab2018study, parthasarathy2018ladder, zhang2019exploiting, tzirakis2017end, ringeval2017summary, chang2017learning}, most often by considering expression of valence (positive vs negative) and arousal (calm vs excited)~\cite{busso2013toward, cowie2001emotion, cowie2003describing}. These dimensional labels can be obtained by asking human raters to annotate data either statically, over units of input (e.g., segments or utterances)~\cite{parthasarathy2017jointly, zhang2017cross}, or continuously in time with a fixed sampling rate~\cite{han2017prediction, huang2017continuous}.

Continuous assessments have the advantage of providing fine-grained information about the emotional state of the speaker as a function of time. 
One of the major challenges in recognizing continuous emotion labels (continuous emotion recognition) stems from annotators' reaction delay, which is defined as the amount of time it takes for annotators to sense acoustic events, understand them, and report the emotional labels~\cite{huang2015investigation, mariooryad2015correcting, nicolle2012robust}. Reaction delay is a convolutive noise that can shift continuous emotion annotations forward in time, causing a time difference between speech signals and the continuous emotion labels. This time difference depends on affective behaviors~\cite{mariooryad2015correcting}, which makes continuous emotion recognition challenging. Therefore, in order to design an effective continuous emotion recognizer, it is crucial to compensate for the reaction delays.

The measure of human reaction delay and its relevance began in the modern era with the emerging focus on detailed experimental calculations in astronomy~\cite{boring1950history}. It was quickly established that reaction time is \textit{individual dependent}~\cite{mollon1996errors}, \textit{stimulus dependent}~\cite{nicolas1997speed}, and \textit{task dependent}~\cite{mariooryad2015correcting, mariooryad2013analysis}. 

The importance of reaction delay compensation in continuous emotion recognition is widely recognized ~\cite{huang2015investigation, mariooryad2015correcting, mariooryad2013analysis, nicolaou2010automatic, trigeorgis2016adieu, nicolaou2011continuous, nicolle2012robust}. There are two main approaches for handling this delay: (1) explicit compensation, in which researchers remove the delay in advance and then train emotion recognition systems~\cite{mariooryad2015correcting, mariooryad2013analysis,nicolaou2010automatic,trigeorgis2016adieu, huang2015investigation}; these approaches assume that the delay compensation and the emotion recognition are independent; (2) implicit compensation, in which researchers build classifiers with large numbers of parameters to compensate for delay~\cite{khorram2017capturing, ringeval2015prediction, le2017discretized}. In this paper, we introduce a continuous emotion recognition system that is able to accomplish both goals: it compensates for annotators' delays while modeling the relationship between speech features and emotion annotations using classifiers with smaller numbers of parameters.


The proposed system is a convolutional network that is able to directly learn the annotators' delays. The network contains two components: (1) an emotion predictor and (2) an aligner. We train both components simultaneously in an end-to-end manner. The emotion predictor is a common convolutional network that models the relationship between acoustic features and emotion labels. The aligner network compensates for the annotators' delays using a new layer, the \textit{delayed sinc layer}, which applies a learnable time-shift to a signal. The delayed sinc layer can be added to any network to compensate for misalignments between two signals. 

The delayed sinc layer takes a one-dimensional signal and passes it through a shifted low-pass (sinc) filter. The layer modifies its input by introducing a fixed delay, the amount of which is trainable through the back-propagation algorithm. We can handle variable delays by incorporating multiple parallel delayed sinc layers, each operating on a different region of the acoustic space.
The shifted low-pass filter is also able to remove high frequency components of the input and generate a smooth output that is consistent with the slow moving nature of the emotion labels obtained from annotators (the importance of which is discussed in~\cite{huang2015investigation}).

The novelty of this work is the introduction and evaluation of the delayed sinc layer.  We evaluate the proposed system on two publicly available continuously annotated emotion corpora: RECOLA~\cite{ringeval2013introducing} and SEWA~\cite{ringeval2017avec}.  We find that the proposed architecture obtains audio-only state-of-the-art performance on RECOLA.  It also obtains audio-only state-of-the-art performance on SEWA when fused with another existing approach~\cite{chen2017multimodal}.  We demonstrate that a system that explicitly compensates for variable delay can use fewer parameters than one that implicitly compensates for delay through a large receptive field.  We further demonstrate that a system that allows for flexibility in delay compensation outperforms systems that do not have this flexibility, noting that a single delayed sinc layer is not enough for modeling annotators' delay because this delay is non-constant. We find that arousal and valence prediction need at least 8 and 16 components, respectively, and delays over 7.5-seconds do not contribute to system performance. This suggests the importance of considering variability in delay (otherwise, the ideal number of components would be one).  Finally, we investigate how reaction lag changes based on laughter, an emotionally salient event.  We find that laughter regions of speech require smaller delays to be aligned with their emotion labels, compared to non-laughter regions.  



\section{Background}

As the main focus of this paper is to model and compensate for annotators' reaction delays, we provide a detailed overview for reaction delay compensation methods in Section~\ref{sec:Reaction Delay Compensation}. We also explain the state-of-the-art continuous emotion recognition systems developed for different datasets in Section~\ref{sec:Continuous Emotion Recognition}. We compare our proposed network with two of these state-of-the-art systems in our experiments.

\subsection{Reaction Delay Compensation}\label{sec:Reaction Delay Compensation}
Reaction delay compensation techniques can be categorized into 2 groups: explicit and implicit.

\subsubsection{Explicit Compensation}
In this approach, the delay compensation and the emotion prediction are performed separately, which removes the need for the emotion predictor to handle delay. Researchers estimate the reaction delays by optimizing an alignment measure through a search algorithm. Different alignment measures have been proposed in different systems including mutual information and emotion recognition performance: 

\textbf{\textit{Mutual information --}} Mariooryad et al. estimated evaluator delay by maximizing the mutual information between the acoustic events and the continuous emotional labels. Their experiments show that the mutual-information-based delay compensation technique can lead to seven percent relative accuracy improvement over baseline classifiers on the SEMAINE dataset~\cite{mariooryad2015correcting, mariooryad2013analysis}.


\textbf{\textit{Recognition performance --}} 
The most common measure is the accuracy of the emotion recognition system. Trigeorgis et al. estimated the reaction time as a fixed value (between 0.04 and 10 seconds) that could be found through maximizing the concordance correlation coefficient between real and predicted emotion labels~\cite{trigeorgis2016adieu}. Huang et al. studied the effect of annotator's delay and introduced a number of multimodal emotion recognition systems based on an output associative fusion technique. They applied a temporal shift to each training sample to compensate for the annotation delay~\cite{huang2015investigation}. This temporal shift is performed by dropping first $N$ emotion labels and last $N$ input features. The value of the temporal shift, $N$, is tuned based on the development error during the training procedure. Their experimental results on AVEC 2015 confirm the importance of the delay compensation for continuous emotion recognition systems. They found that the best delays for arousal and valence are four and two seconds, respectively.

However, these approaches assume that the reaction delay is fixed for different acoustic events and that delay compensation and emotion prediction are independent and can be trained separately. 

\subsubsection{Implicit Compensation}
In this approach, researchers leverage models that are able to compensate for delays while modeling the relationship between speech features and emotion labels. Different models have been used to this, including LSTM and convolutional networks:

\textbf{\textit{LSTM network}} -- 
Ringeval et al.~\cite{ringeval2015prediction} applied a long short-term memory (LSTM) network~\cite{sak2014long} with an analysis window to deal with simultaneous modeling of reaction delays and emotion labels. They found that predicting valence requires a longer analysis window compared to predicting arousal. Le et al. applied a multi-task bidirectional (B)LSTM network to model continuous emotion labels in a categorical time-dependent framework~\cite{le2017discretized}. Their network is trained with a cost-sensitive cross-entropy loss function. 

\textbf{\textit{Convolutional network}} --
In our previous paper~\cite{khorram2017capturing}, we employed two convolutional networks based on dilated convolutions~\cite{yu2015multi} and downsampling/upsampling layers~\cite{noh2015learning, badrinarayanan2015segnet} to predict continuous emotion labels. Both networks have large receptive fields that allow the networks to automatically shift the inputs forward in time and compensate for the reaction delay. 

However, these implicit modeling approaches are not specifically designed to compensate for reaction delays. In this paper, we introduce the delayed sinc kernel, which is specifically designed to deal with delays in neural networks. Our experiments show that the proposed structure outperforms previous continuous emotion recognition systems.

\subsection{Continuous Emotion Recognition}\label{sec:Continuous Emotion Recognition}

Continuous prediction of dimensional attributes has gained increasing attention over the last few years~\cite{ringeval2018avec, wang2018towards, wataraka2018speech, zhao2018multi}. Several competitions have been held in this research area, and different continuous emotion recognition systems have been proposed for each competition. For example, the audio/visual emotion challenge (AVEC) is a series of competition events aimed at comparing multimodal methods for recognizing emotion and depression patterns from audio, video, and physiological signals. In this section, we describe state-of-the-art emotion recognition systems developed for the AVEC competitions.


\textbf{\textit{AVEC 2015 -- }}
In the winning submission of the AVEC 2015, He et al.~\cite{he2015multimodal} introduced an emotion prediction system with two phases.  In the first phase, they obtained a set of initial predictions from each input modality using a BLSTM network.  In the second phase, the initial predictions were smoothed with a Gaussian smoothing filter, and input into a multimodal BLSTM network for the final prediction of the affective states.  The authors extracted a comprehensive set of $4,684$-dimensional low-level feature vectors from speech with the frame rate of 25 frames/s using both openSMILE~\cite{eyben2013recent} and YAAFE~\cite{mathieu2010yaafe} toolkits, including loudness, zero crossing rate, spectral flux, Mel-frequency cepstral coefficients (MFCCs), and voicing related features (such as jitter, shimmer, logarithmic Harmonics-to-Noise Ratio (logHNR), etc).  In our previous paper~\cite{khorram2017capturing}, we showed that 
a much smaller $40$-dimensional MFB features with convolutional networks could be used to obtain state-of-the-art performance.  

\textbf{\textit{AVEC 2016 -- }}
In the best system submitted to the AVEC 2016 challenge, Brady et al.~\cite{brady2016multi} employed a set of low-level audio features including MFCCs, shifted delta cepstral, and prosody features.  The authors then trained a sparse coding technique over the low-level features to extract a set of higher-level audio features.  The resulting higher-level features were then used to train linear SVRs for final continuous emotion recognition.  In another successful system, Povolny et al.~\cite{povolny2016multimodal} extracted two sets of low-level audio features:
(1) extended Geneva minimalistic acoustic parameter set (eGeMAPS)~\cite{eyben2016geneva} and (2) bottleneck features obtained from intermediate representations of a DNN trained for automatic speech recognition (ASR) application.  This DNN-based ASR is trained over an initial set of $24$ log Mel filterbank (MFB) features and four different estimates of the fundamental frequency (F0).  Povolny et al. used two methods for combining low-level features into high-level features that capture local contextual information: (1) simple frame stacking and (2) temporal content summarization by calculating statistics over local windows.  The authors trained linear regressors to generate emotional labels from high-level statistics.  

In our previous work~\cite{khorram2017capturing}, we showed that capturing long-term dependencies is beneficial for continuous emotion recognition.  We studied two CNN-based architectures that are able to capture long-term temporal dependencies in a given sequence of acoustic features: dilated CNN and downsampling/upsampling network.  Dilated CNN uses a stack of convolutional kernels with varying dilation factors to capture long-term temporal dependencies.  We showed that the dilated CNN outperforms previous systems, but it has an important problem: the output signals generated from this network undergo irregular changes between successive time steps.  This noisy output is not consistent with the slow-moving emotion labels that are defined by annotators.  
We showed that a downsampling/upsampling network could be used to generate a smooth signal while considering the long-term dependencies for predicting emotion.  
It applies a series of convolutions and max-poolings to compress the input signal into a downsampled (low-resolution) signal.  It then applies a series of transposed convolution\footnote{In the literature transposed convolution is also known as deconvolution, upconvolution, fractionally strided convolution and backward strided convolution.}~\cite{zeiler2010deconvolutional, dumoulin2016guide} layers to upsample the compressed signal and generate the target emotional labels.  This network achieved the best audio-only performance on the AVEC 2016 challenge.  In this paper, we demonstrate that the proposed multi-delay sinc network outperforms the downsampling/upsampling network on the AVEC 2016 data.

\textbf{\textit{AVEC 2017 -- }}
In the winning system of the AVEC 2017~\cite{chen2017multimodal}, Chen et al. implemented a multi-task system that models and predicts multiple emotion labels simultaneously.  They used IS10~\cite{schuller2011recognising} and Soundnet features (intermediate representations of the pretrained Soundnet network~\cite{aytar2016soundnet}) to train an LSTM-based continuous emotion predictor.  LSTM network is able to alleviate the annotation delay problem and reduce the feature engineering efforts.  In this paper, we show that fusing the predictions of the proposed network and the IS10-based system~\cite{chen2017multimodal} yields the best audio-only performance on the AVEC 2017.




\section{Datasets and Features}\label{Dataset and Features}

In this section, we introduce the datasets, the features, the metrics, and the evaluation schemes used in this paper.  We use two evaluation schemes, one based on the AVEC challenge (Section~\ref{sec: AVEC scheme}) and one using leave-one-speaker-out cross-validation scheme (Section~\ref{sec:loso}).

\subsection{Datasets}\label{Datasets}
We use two publicly available datasets to conduct the experiments of this paper: (1) the remote collaborative and affective interactions (\emph{RECOLA}) dataset~\cite{ringeval2013introducing} and (2) a subset of sentiment analysis in the wild (\emph{SEWA})\footnote{http://sewaproject.eu} dataset.  Both provide audio-visual recordings that capture spontaneous and naturalistic behaviors of subjects and are annotated with continuous emotion labels (arousal and valence values). 

\subsubsection{RECOLA}
RECOLA was used in the multimodal affect recognition sub-challenge of AVEC 2016.  It contains 27 samples of spontaneous and naturalistic interactions that were collected from 27 different French-speaking subjects.  All samples are five minutes in length and include audio, video, electro-cardiogram (ECG) and electro-dermal activity (EDA).  In this work, we focus only on the audio modality. The samples were partitioned uniformly into train, development and test sets by the organizers, nine per set.

The data were evaluated by six annotators (three female and three male), sampled at 25Hz.  The evaluations include continuous assessments of valence and arousal.  The six evaluation traces are fused into a single ground-truth using the protocols described in the AVEC 2016 challenge~\cite{valstar2016avec}.  Emotion labels are available only for the training and development sets of the data.  Performance on the testing data is assessed by the organizers of the AVEC 2016 challenge.

\subsubsection{SEWA}
SEWA was used in the affect recognition sub-challenge of AVEC 2017.  It is an audio-visual dataset of human-human interactions collected using common web-cams and microphones over the OpenTok API\footnote{https://tokbox.com}, an online platform for setting up a video call.  Each recording includes two subjects discussing arbitrary aspects of a commercial that they have just viewed.  The conversations range in length from 47-seconds to 3-minutes.  

SEWA is a multi-lingual dataset, but the affect recognition challenge of AVEC 2017 used only the German-language recordings~\cite{ringeval2017avec}.  We also use the German-language subset in this paper.  The subset contains 32 dyadic conversations (64 subjects in total), divided into three partitions (34 train, 14 development and 16 test).  The data is split such that both subjects from a recording are in the same partition.

The data were evaluated by six annotators (three female and three male), sampled at 10 Hz.  Again, the evaluations include continuous assessments of valence and arousal.  We use the single gold standard label provided by the challenge~\cite{ringeval2017avec}.  As in RECOLA, the test labels are not released and performance of the system is assessed by organizers of the AVEC 2017 challenge.


\subsubsection{Differences between RECOLA and SEWA}\label{dataset_differences}
There are three important differences between the RECOLA and the SEWA datasets that should be considered in the design of our models:\vspace{0.1cm}
\begin{enumerate}
  \item In SEWA, each recording contains a conversation between two partners, one ``target'' and the other ``non-target''. The target partner is the partner whose emotions we aim to predict. In RECOLA, although the data were obtained from dyadic conversations, each recording includes only the audio from the target speaker.\vspace{0.1cm}
  \item In RECOLA, each recording has the same duration (5-minutes).  In SEWA, the duration varies from 47-seconds to 3-minutes.\vspace{0.1cm}
  \item In RECOLA, the sampling rate of the emotion annotation traces is 25Hz.  In SEWA, it is 10Hz.
\end{enumerate}

\subsection{Evaluation}\label{Evaluation}
The AVEC 2016 and 2017 challenges use the root mean square error (RMSE) and concordance correlation coefficient (CCC) metrics.  RMSE is the standard error metric.  CCC measures the agreement between two signals.  It ranges from -1 to 1 and it is zero when two signals are uncorrelated from each other. It is defined by:

  $$\text{CCC} = \frac{2\sigma_{y\hat{y}}}
   			   {\sigma^{2}_{y} + \sigma^{2}_{\hat{y}} + 
               (\mu_{y}-\mu_{\hat{y}})^2}$$
\\
where $y$ and $\hat{y}$ are the sequences of ground-truth and predicted labels;
$\mu_{y}$ and $\mu_{\hat{y}}$ are the mean of $y$ and $\hat{y}$;
$\sigma^{2}_{y}$ and $\sigma^{2}_{\hat{y}}$ are the variance of $y$ and $\hat{y}$; $\sigma^2_{y\hat{y}}$ is the covariance between $y$ and $\hat{y}$.



\subsubsection{AVEC evaluation scheme}\label{sec: AVEC scheme}
This scheme follows the AVEC 2016 and 2017 guidelines.  We train systems on the training/development partitions and assess performance on the test portion.  We concatenate the output from each test recording into a single vector, which is then used to calculate the RMSE and CCC evaluation metrics.  It is important to note that statistical tests cannot be performed in this setting because a single value is computed over all speakers and there are limited numbers of submissions allowed, thus precluding repeated assessments.  

We train the network on the training partition, optimizing using the CCC metric over different sets of tuning parameters.  We use the development partition to identify the set of tuning parameters that result in the highest performance. Finally, we use the identified network to generate labels for the held-out test data and submit the predictions to the organizers of the challenge.  The organizers compute the final test evaluation metrics.


\subsubsection{Leave-one-speaker-out evaluation scheme}
\label{sec:loso}
The leave-one-speaker-out scheme addresses the limitation introduced by the AVEC challenge guidelines: the lack of ability to assess statistical significance.  In this scheme, we perform leave-one-speaker-out cross-validation over the development speakers.  We first train multiple networks with different hyper-parameters by maximizing average CCC values of the training samples. We then select hyper-parameters in a leave-one-speaker-out manner over the development set.  We split the development set into speaker-specific folds (9 folds for AVEC 2016 and 16 folds for AVEC 2017).  We leave out one fold for testing and choose the hyper-parameters using the remaining folds.  We calculate metrics for each left out fold separately and then report the average as the final performance.

\subsection{Features}\label{Features}
Speech processing systems have relied upon a diverse set of spectral features, including linear prediction coefficients (LPC)~\cite{eyben2009openear}, perceptual linear prediction (PLP) coefficients~\cite{eyben2009openear}, mel-frequency cepstral coefficients (MFCC)~\cite{muda2010voice, khorram2016recognition}, mel-generalized cepstral coefficients (MGC)~\cite{tokuda1994mel, khorram2014context, khorram2015soft, khorram2013speech}, and log mel-frequency bank (MFB) features~\cite{busso2007using, khorram2018priori}.  Emerging work has shown that emotion recognition systems can effectively use feature vectors composed solely of MFB features ~\cite{le2017discretized, aldeneh2017pooling} and that this small feature set can outperform much larger feature sets~\cite{khorram2017capturing}.  In this work, we extract 40-dimensional MFB features using the Kaldi toolkit~\cite{povey2011kaldi} with a $25$ms Hann window and $10$ms frame shift.  We apply speaker-specific $z$-normalization to reduce the speaker variability in the extracted features.

Feature vectors for the RECOLA dataset are created by concatenating every four successive MFB vectors to form a 160-dimensional vector;  this creates a feature vector sampling rate that is consistent with the sampling rate of emotional labels~\cite{khorram2017capturing, le2013emotion}.  The SEWA dataset is more complicated to process because the dataset contains speech from both target and non-target speakers.  We follow a method, introduced by Chen et al.~\cite{chen2017multimodal}, that creates an 80-dimensional vector (40-dimensions per speaker).  If a frame contains speech from the target speaker, the first half of the vector takes MFB values and the second half is zero and vice versa.  We then concatenate every 10 consecutive feature vectors to again make the feature vector sampling rate consistent with the sampling frequency of emotion labels (10Hz).  This process results in a sequence of 800-dimensional acoustic features.

\section{Preliminary Experiment}\label{Preliminary Experiment}
In this section, we set up an experiment that searches for an effective delay for RECOLA and SEWA in a manner similar to~\cite{mariooryad2015correcting}.  In later sections we will demonstrate that our networks are capable of learning these delays (see Section~\ref{sec: learning delay through delayed sinc layer}).  The goal is to explicitly compensate for delay.  We first time-shift all speech signals using a static shift; we then train a network that recognizes emotion labels from the delayed speech signals; finally, we calculate the leave-one-speaker-out CCC of the trained network; we repeat this procedure for different values of delay ranging from 0 to 6 seconds (step size of 400 milliseconds), and study the effect of the delay on the CCC results. 

We employ a neural network consisting of a convolutional layer with one filter of length 2 seconds (50 frames on RECOLA and 20 frames on SEWA), followed by a $tanh$ activation unit, followed by a linear regression layer. The network is trained using the Adam optimizer over the CCC metric~\cite{kingma2014adam}. We select the number of training epochs and calculate the CCC values using the leave-one-speaker-out evaluation scheme explained in Section~\ref{Evaluation}.

\begin{figure}[t]
\centering
\def\factor{0.7}
\includegraphics[width=\factor\linewidth]{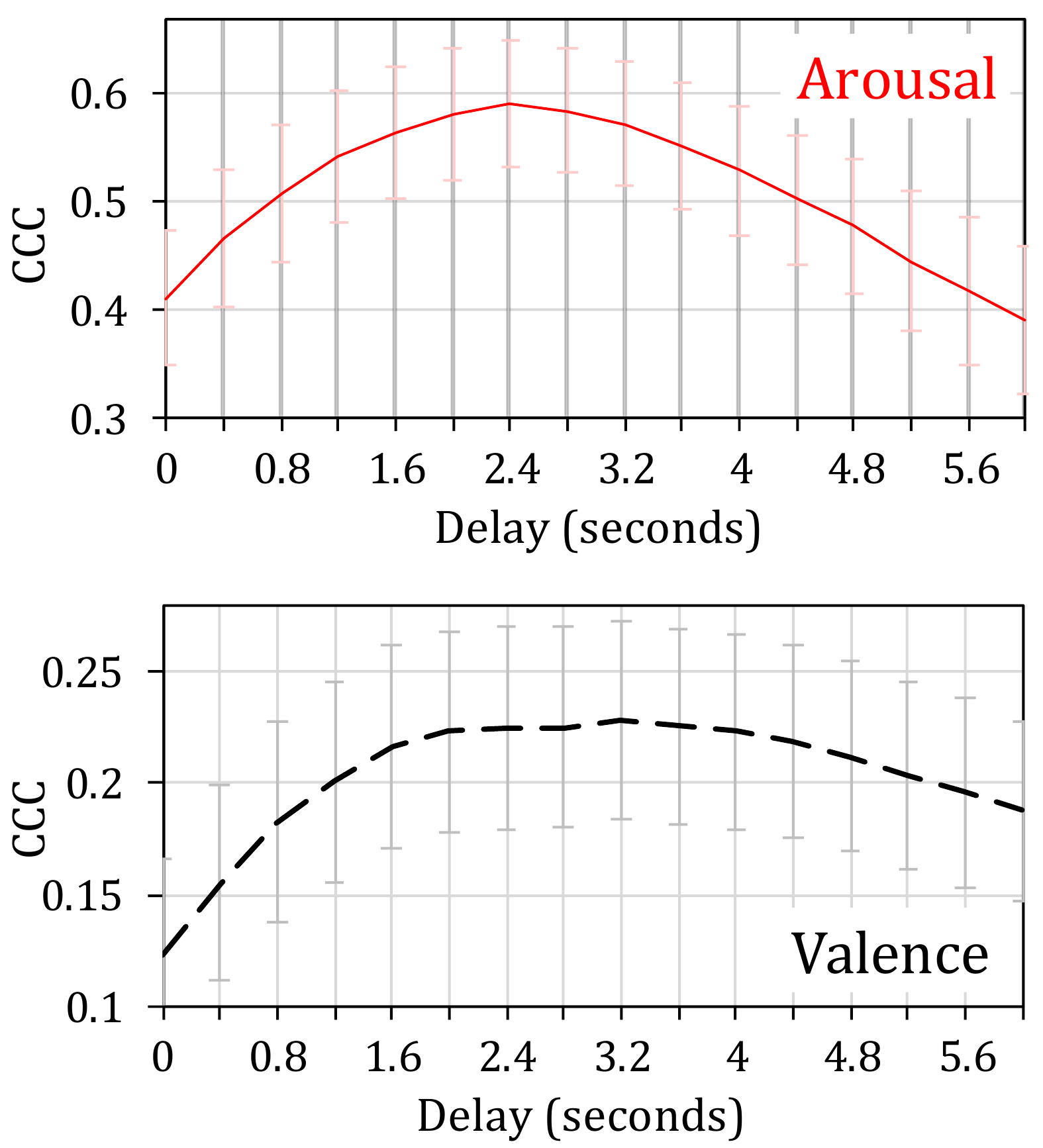}
\\\hspace{0.8cm}Results of the \textbf{RECOLA} dataset\vspace{0.4cm}
\includegraphics[width=\factor\linewidth]{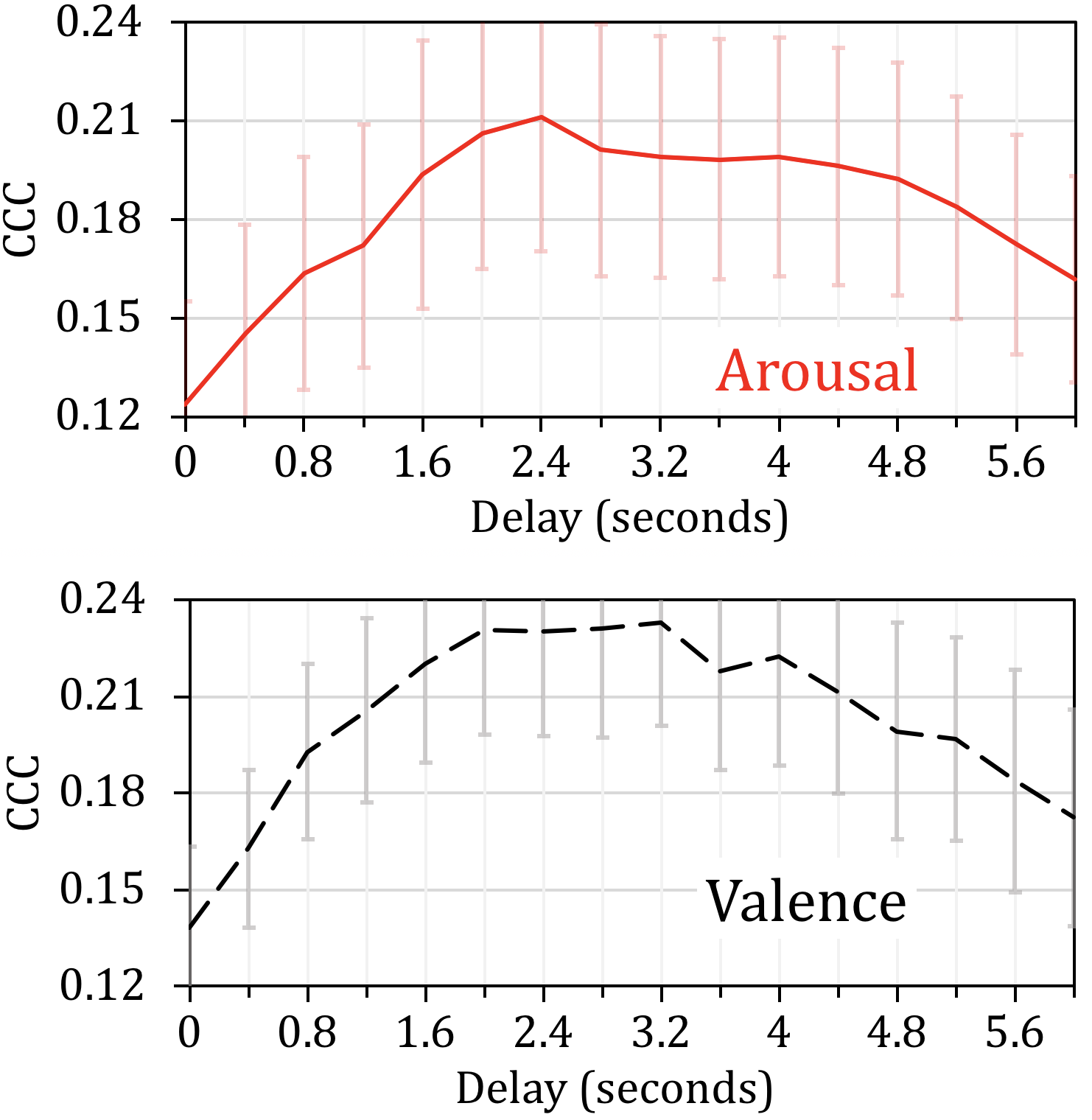}
\\\hspace{0.9cm}Results of the \textbf{SEWA} dataset
\caption{Applying delay to acoustic features improves mean CCC for both arousal (solid red curve) and valence (dashed black curve). Error bars show standard deviation across different subjects in leave-one-speaker-out evaluation.\vspace{-0.2cm}}
\label{fig:preliminary_experiment}
\end{figure}

Figure~\ref{fig:preliminary_experiment} shows the CCC results with respect to the delay values. The results show that increasing the delay up to 2.4 seconds for arousal and 2 seconds for valence improves the leave-one-speaker-out CCC.
It shows that synchronizing input and output (compensating for annotators' delay) is important in continuous emotion recognition. 
This result agrees with the previous findings (reported in~\cite{mariooryad2015correcting}). We also find that when predicting valence, performance does not change given delays in the range of 2 to 3.2 seconds. This shows that we cannot find a unique value as the best estimate of the annotators' delay in detecting valence (i.e., the delay process follows a multi-modal distribution), which motivates the need to apply multiple delays for predicting continuous emotion labels.

This analysis presupposes that the reaction lag is a parameter of the network that can be tuned using validation data. This method of synchronizing input and output has several problems: (1) a separate network must be trained for any candidate value of the delay which is resource intensive; (2) the estimated lag must be a multiple of the sampling rate; (3) using this method, we cannot apply different delays to different regions of the acoustic features. In the next section, we introduce a method that can solve these problems.


\section{Methods}\label{Methods}
In this section, we introduce the \emph{delayed sinc layer}: a convolutional layer that is able to apply a learnable shift to a given input signal.  The delayed sinc layer is a time-shifted low-pass filter: it passes the frequency components that are lower than a certain cutoff frequency and introduces a unique time-shift to its input. It allows us to align two signals in a neural network architecture.


We also demonstrate how the delayed sinc layer can be used in a continuous emotion recognition system. Our final network uses multiple delayed sinc layers and fuses the final outputs to compensate for signals that have multiple or time-varying delays.

\subsection{Delayed Sinc Layer}\label{Delayed sinc Layer}
Let $x(t)$ and $y(t)$ be continuous signals that are defined for acoustic features and emotion labels, at time $t$. The goal of continuous emotion recognition is to find a mapping $\mathcal{F}_{cer}$ that takes a sequence of acoustic features up to time $t$, $X(t)=[x(t')\ \forall\ t' \leq t]$, and estimates its corresponding emotion label at time $t$, $y(t)$: 
\begin{equation} \label{eq_fcer}
y(t) = \mathcal{F}_{cer}(X(t)).
\end{equation}
According to the preliminary experiment reported in Section~\ref{Preliminary Experiment}, $x(t)$ and $y(t)$ are not synchronized. Therefore, $\mathcal{F}_{cer}$ should be written based on two simpler mappings: $\mathcal{F}_{syn}$ and $\mathcal{F}_{pred}$. $\mathcal{F}_{syn}$ performs the synchronization and $\mathcal{F}_{pred}$ models the relationship between acoustic features and the synchronized labels. In other words: 
\begin{equation} \label{eq_pred}
y(t) = \mathcal{F}_{cer}(X(t)) = \mathcal{F}_{syn}(\mathcal{F}_{pred}(X(t))).
\end{equation}

For the sake of simplicity, this section assumes that the synchronization can be done through applying a fixed delay $\tau$ (we will relax this assumption in Section~\ref{Multi-Delay sinc Network}). In this case, $\mathcal{F}_{syn}$ can be easily implemented through convolving $\mathcal{F}_{pred}(X(t))$ with a time-shifted dirac-delta function. Therefore, $y(t)$ can be written as:
\begin{equation} \label{eq_syn_written}
y(t) =  \mathcal{F}_{syn}(\mathcal{F}_{pred}(X(t))) = \mathcal{F}_{pred}(X(t)) * \delta(t-\tau),
\end{equation}
where $*$ and $\delta$ represent the convolution operator and the dirac-delta function, respectively.

An effective approach to estimate $\tau$ is to learn it along with the parameters of $\mathcal{F}_{pred}$ through a gradient-based optimization technique. However, $\delta(t)$ is not a differentiable function whenever $t=0$. Therefore, $\tau$ as a parameter of $\delta(t-\tau)$, is not directly learnable in this manner. To solve this problem, we approximate the $\delta$ function with a smoother function. Below, we show that the sinc function is an appropriate approximation of the dirac-delta function for generating continuous emotion curves; sinc function generates smooth curves that are consistent with the slow-moving ground-truth curves of emotions.  

Another issue is that the mapping function, $\mathcal{F}_{pred}$, may generate a signal that contains high-frequency components, which is not desirable in continuous emotion recognition because human annotations are smooth and slow-moving. In our previous paper, we showed that incorporating a temporal smoothing technique into the network architecture can improve the performance~\cite{khorram2017capturing}. We propose to apply a low-pass filter, $h_{lp}(t)$, to the signal generated by $\mathcal{F}_{pred}$, i.e.,
\begin{equation} \label{eq_syn_lpf}
y(t) =  (\mathcal{F}_{pred}(X(t)) * h_{lp}(t)) * \delta(t-\tau),
\end{equation}
which is equivalent to
\begin{equation} \label{eq_syn_lpf_simple}
y(t) =  \mathcal{F}_{pred}(X(t)) * h_{lp}(t-\tau).
\end{equation}
Accordingly, we can use a time-shifted low-pass filter instead of a dirac-delta function to compensate for  reaction lag and also remove the unsatisfactory high-frequency components from the output. An ideal low-pass filter, $h_{lp}(t)$, is the $sinc$ function, which can be expressed by:
\begin{align}\label{eq_sinc}
\begin{split}
h_{lp}(t) = 2f_c\ sinc(2{f_{c}}t),\hspace{0.36cm}\\[0.1cm]
sinc(t)=
\left\{\begin{matrix}
\frac{\sin(\pi t)}{\pi t} & t \neq 0 \vspace{2mm} \\ 
1 & t = 0
\end{matrix},\right.
\end{split}
\end{align}
where $f_{c}$ is the cutoff frequency (also known as bandwidth), which is defined as the maximum frequency that the sinc filter does not attenuate. 

The cutoff frequency of the sinc filter, $f_{c}$, must be higher than the maximum frequency of the ground-truth signal, $f_{g}$, (i.e., $f_{c} \geqslant f_{g}$). Otherwise, the sinc filter cannot pass all frequencies of the predicted emotional labels and the sinc output will be smoother than the actual ground-truth labels. 

For a real discrete-time signal, the frequency components ranges from $0$ to $\frac{f_s}{2}$, where $f_s$ is the sampling frequency. In this case, the sinc filter with $\frac{f_s}{2}$ cutoff frequency passes all frequencies without attenuating them. Therefore, sinc filter with $\frac{f_s}{2}$ cutoff frequency can be used to apply a delay to any real signal sampled at $f_s$.

The sinc filter expressed by equation~(\ref{eq_sinc}) has infinite number of coefficients and therefore it is not implementable in practice.
In order to approximate it, a windowed-sinc filter is commonly used instead of the ideal low-pass filter~\cite{mitra2006digital, khorram2019trainable}. Equation~(\ref{eq_inout_windowed}) expresses the input-output relationship after applying a window $h_w(t)$ to the sinc filter:
\begin{equation} \label{eq_inout_windowed}
    y(t) = \mathcal{F}_{pred}(X(t)) * \Big(2{f_c}\ sinc(2f_{c}(t-\tau))\ h_w(t)\Big)
\end{equation}
Applying $h_w(t)$ causes distortion to the ideal frequency response of the sinc filter. In our initial experiments, we noticed that the type of the window does not significantly change the final predictions; therefore, we employ a simple rectangular window in our experiments. Equation~(\ref{eq_inout_windowed}) expresses our convolutional layer in the continuous time domain. To implement it, we must discretize Equation~(\ref{eq_inout_windowed}) using the sampling frequency of $f_s$ (25Hz for RECOLA and 10Hz for SEWA), which leads to the following convolutional kernel for our delayed sinc layer: \begin{equation} \label{eq_sinc_kernel}
    h_{sinc}[n; \tau]=(2f_c/f_s)\ sinc[2f_{c}(n/f_s-\tau)]\ h_w[n].
\end{equation}

In summary, the delayed sinc layer is a convolutional layer that uses a special kernel. The shape of the kernel is limited to a time-shifted sinc, expressed by equation~(\ref{eq_sinc_kernel}). Delayed sinc layer has 3 parameters:\vspace{0.1cm}

\begin{enumerate}
    \item Delay parameter, $\tau$: Delayed sinc layer introduces a delay of $\tau$ seconds to its input. $\tau$ is the only parameter of the delayed sinc layer that has to be trained. In our experiments, we initialize $\tau$ randomly using a uniform distribution between 0 and 20 seconds.\vspace{0.1cm}
    \item Bandwidth of the sinc kernel, $f_{c}$: It is a constant parameter that must be higher than the bandwidth of the ground-truth signals. We set this parameter to $\frac{f_s}{2}$ in our experiments, where $f_s$ is the sampling frequency of the ground-truth signals (i.e., 25Hz for the RECOLA dataset and 10Hz the SEWA dataset)\vspace{0.1cm}
    \item Windowing function, $h_w[n]$: We use a long rectangular window with the length of 44 seconds to be sure that the window does not remove the main beam of the sinc function even after applying the longest initial delay (i.e., 20 seconds). Additionally, this window results in a network with 44 seconds receptive field which is consistent with the effective receptive field found in~\cite{khorram2017capturing}.
\end{enumerate}

\section{Multi-Delay Sinc Network}\label{Multi-Delay sinc Network}

In the previous section, we introduced a convolutional kernel that can compensate for a time-invariant delay. Delays introduced by human annotators, however, are not necessarily constant in time; annotators may introduce different delays to different regions in the input.  For example, it is easier to identify emotions for laughter parts of speech~\cite{truong2005automatic, bickley1992acoustic} and therefore annotators may be able to identify them faster. This section introduces a new network architecture, the \emph{multi-delay sinc (MDS)} network, that utilizes multiple delayed sinc layers to deal with time-variant reaction delays.

We first describe how multiple delayed sinc layers are used.  Each delayed sinc layer is applied to a different region of the acoustic space, formed by generating fuzzy clusters.  We then describe a network architecture that can integrate the prediction from multiple layers.

\subsection{Acoustic Clustering}
\label{sec:cluster}
\begin{figure}[t]
    \centering
    \def\factor{0.95}
    \includegraphics[width=\factor\linewidth]
    {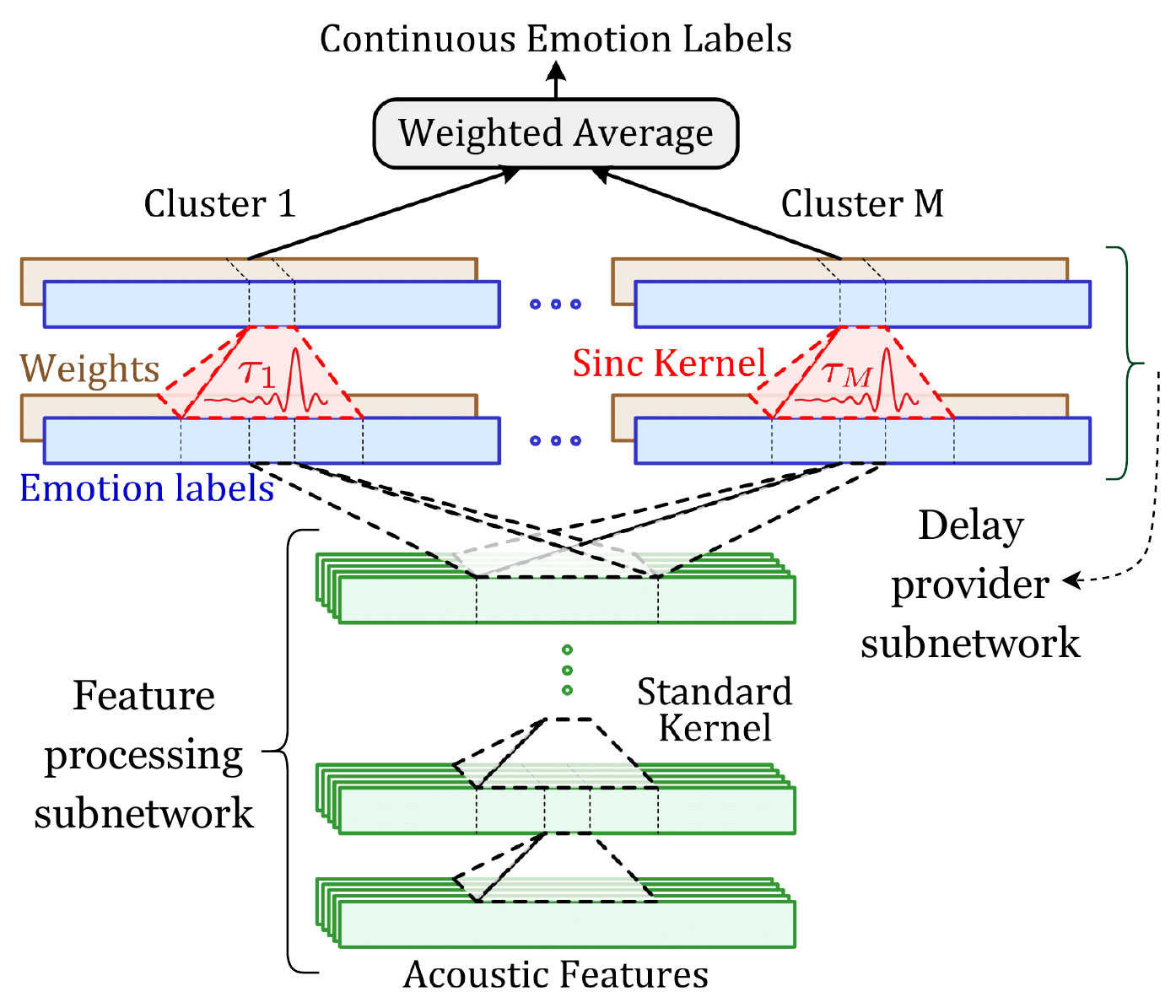}
    \caption{A visualization of our multi-delay sinc network with $M$ clusters. $\tau_m$ is the delay considered for the $m$-th component. Standard and sinc kernels are shown in black and red colors, respectively. In this figure, we show the structure of the first and the last clusters.\vspace{-0.2cm}}
    \label{fig:system_block_diagram}
\end{figure}

Figure~\ref{fig:system_block_diagram} shows the architecture of the MDS network with $M$ clusters. The main idea of the MDS network is to categorize speech regions into a number of fuzzy clusters such that all samples associated with a cluster require the same delay to be synchronized with the ground-truth labels.  The system is trained in an end-to-end manner and fuzzy membership functions of clusters are implicitly learned.  The number of clusters, $M$, is a parameter that must be tuned. More precisely, the MDS network defines three components for each cluster $m$: \vspace{0.1cm} 
 
 \begin{enumerate}
     \item $\tau_m$: a learnable delay which is a single parameter for each cluster. $\tau_m$ will be trained along with other parameters of the network. \vspace{0.1cm} 
     \item $\mathcal{F}_m$: emotion recognition, a mapping that predicts emotion labels for the $m$-th cluster based on input features, $X$. $\mathcal{F}_m[n; X]$ denotes the $n$-th sample of the signal generated by the mapping $\mathcal{F}_m$. We employ a standard multi-layer convolutional neural network to generate $\mathcal{F}_m[n; X]$ from $X$.\vspace{0.1cm} 
     \item $w_m$: a mapping that generates a weight signal for the $m$-th cluster using input features, $X$. $w_m[n; X]$ denotes the $n$-th sample of the signal generated by the $w_m$ mapping. $w_m[n; X]$ quantifies the importance of incorporating $\mathcal{F}_m[n; X]$ into the final predictions $y[n]$. A standard convolutional neural network is employed to define the mapping $w_m$.\vspace{0.1cm} 
 \end{enumerate}

\subsection{Network Architecture}
The proposed MDS network takes MFB features as input and passes them through a stack of three subnetworks: (1) feature processing, (2) delay provider, and (3) averaging subnetworks. The subnetworks are shown in Figure~\ref{fig:system_block_diagram}. \vspace{0.1cm} 

\subsubsection{Feature processing subnetwork}

The feature processing subnetwork takes acoustic features and generates emotion labels, $\mathcal{F}_m[n; X]$, and weight signals, $w_m[n; X]$ for all clusters. The MDS network uses a shared multilayer convolutional network to simultaneously generate all the emotion labels ($\mathcal{F}_m[n; X]$) and weight signals ($w_m[n; X]$). Our initial experiments showed that using a separate network to generate emotion labels and weight signals does not improve the results. The output of this network is $M$ label signals and $M$ weight signals, where $M$ is the number of clusters. Each label and weight signal is a one-dimensional signal with the same length of the final emotion predictions.

\subsubsection{Delay provider subnetwork}

This subnetwork applies a cluster-specific delay, $\tau_m$, to both labels ($\mathcal{F}_m[n; X]$) and weights ($w_m[n; X]$) using a delayed sinc kernel. Suppose $\vec{\mathcal{F}}_m[n;X,\tau_m]$ and $\vec{w}_m[n;X,\tau_m]$ are the delayed labels and weights predicted for $m$-th cluster. Then,
\begin{equation} \label{eq_delay_provider1}
\vec{\mathcal{F}}_m[n;X,\tau_m] = \mathcal{F}_m[n;X] * h_{sinc}[n;\tau_m],
\end{equation}
\begin{equation} \label{eq_delay_provider2}
\vec{w}_m[n;X,\tau_m] = w_m[n;X] * h_{sinc}[n;\tau_m],
\end{equation}
where $h_{sinc}[n;\tau_m]$ is the windowed sinc kernel expressed by Equation~(\ref{eq_sinc_kernel}). This subnetwork generates a series of predictions that are hypothesized to be more closely aligned with the input features. This subnetwork has $M$ parameters $\{\tau_m\ \forall\ m\in\{1...M\}\}$ that have to be trained.

\subsubsection{Averaging subnetwork}
The previous sections explained how feature processing and delay provider subnetworks generate weights $\vec{w}_m[n;X,\tau_m]$ and emotion labels $\vec{\mathcal{F}}_m[n;X,\tau_m]$ for the $m$-th cluster. The goal of the averaging sub-network is to generate final emotion labels $y[n]$ by combining cluster-dependent labels through cluster weights. One straightforward approach is to use the emotions predicted by the most likely cluster (the cluster with maximum weight); i.e.,
\begin{equation} \label{eq_max_averaging}
    \hat{m}[n] = \arg\max_{m} \vec{w}_m[n;X,\tau_m],
\end{equation}
\begin{equation} \label{eq_max_averaging1}
    y[n] = \vec{\mathcal{F}}_{\hat{m}[n]}[n;X,\tau_{\hat{m}[n]}],
\end{equation}
where $\hat{m}[n]$ is the index of the most likely cluster (cluster with maximum weight) at time n. However, there are two problems with this approach: (1) the assumption that a part of the signal is associated with a single fixed delay is restrictive; and (2)$\hat{m}[n]$ may change at the middle of a recording and as a result predicted labels may experience a sudden change in a recording which is not consistent with the nature of the emotion labels. To deal with this problem, we propose to use a soft-max instead of the standard max. By using soft-max all clusters will contribute in generating the final emotion labels, and therefore it is less likely to observe sudden changes in the final emotion labels. 


In the proposed network, the final continuous emotion label, $y[n]$, is obtained by taking a weighted average of the cluster-specific predictions. The parameters used to weight the predictions are derived from the weights described in Section~\ref{sec:cluster}.  We convert these weights in the previous section to a probability distribution by passing them through a time-distributed softmax layer, which applies a softmax function at each time-step.
Let $\tilde{w}_m[n;X,\tau_m]$ denote the output of the softmax layer, then, 
\begin{equation} \label{eq_softmax_output}
    \tilde{w}_m[n;X,\tau_m] = 
    \frac
    {\exp(\vec{w}_m[n;X,\tau_m])}
    {\sum_{m'=1}^{M}\exp(\vec{w}_{m'}[n;X,\tau_{m'}])}.
\end{equation}

The averaging network uses the cluster probabilities obtained through equation~(\ref{eq_softmax_output}) to calculate the final predictions, $y[n]$:
\begin{equation} \label{eq_final_prediction}
    y[n] = \sum_{m=1}^{M} \tilde{w}_m[n;X,\tau_m]\vec{\mathcal{F}}_m[n;X,\tau_m].
\end{equation}
The averaging subnetwork does not have any parameters to be trained in the training phase. 

We train all these subnetworks in an end-to-end manner using the back-propagation algorithm. The result is the MDS network that automatically assigns features to clusters, applies delay to each cluster, and aggregates the result.

Predicted emotion labels for each cluster ($\vec{\mathcal{F}}_m[n;X,\tau_m]$) are band-limited signals with the maximum frequency of $f_c$ (i.e., cut-off frequency of the sinc filter); however, when we combine them using time-varying weights, the generated continuous emotion labels, $y[n]$, may have higher frequency components. Using multiple clusters enable us to provide more complex delay components, but it has a disadvantage too; it may generate high frequency components in the output, which is not consistent with the slow moving nature of the continuous emotion labels.


\section{Experiments}\label{Experiments}

\subsection{Experimental Setup}\label{Experimental Setup}
We build our models using the TensorFlow numerical computation library~\cite{abadi2016tensorflow}. We train all models by optimizing the CCC metric through the Adam optimizer~\cite{kingma2014adam, gideon2017progressive}. Each network is trained for 300 epochs and the best epoch is selected during validation. To reduce the effect of random initialization, we train each network three times and select the best performing network based on the validation CCC. We use the AVEC and leave-one-speaker-out schemes, explained in Section~\ref{Evaluation}, to calculate validation performance, tune hyper-parameters, and evaluate networks.

\subsubsection{Baseline System}
We implement the RECOLA state-of-the-art audio system, the downsampling/upsampling convolutional network, as the baseline method for comparison, introduced in our prior work~\cite{khorram2017capturing}.  The network first encodes the input signal into a low-resolution signal through a stack of convolution max-pooling layers and then reconstructs the output through a stack of upsampling layers. We exploit transposed convolution layers (deconvolution layers) to upsample the encoded representations and generate the output annotations. We apply the $tanh$ function after each layer (except the final layer which is a linear layer).

We train our downsampling/upsampling network by fixing the learning rate, down-sampling ratio, and number of downsampling layers to 0.0001, 2 and 7, respectively.  We also cross-validate number of kernels (32, 64, 128), length of kernels (3, 4, 5) and L2 regularization weight (0.0, 0.02, 0.04) based on the validation CCC.  We selected these values based on the validation CCC results that we obtained in our initial experiments.

\subsubsection{MDS System}
We implement the proposed MDS system as described in Section~\ref{Multi-Delay sinc Network}.  We apply $tanh$ nonlinearity after all standard convolutional kernels, except the ones that generate cluster weights and labels. We cannot apply any nonlinearity function after delayed sinc layers because sinc kernels have been specifically designed to generate frequency components of the ground-truth labels and applying nonlinearities will change their frequency response. We train the MDS network by fixing the learning rate to 0.001 and cross-validating the number of the standard kernels (16, 32), length of the standard kernels (4, 8, 16), number of the convolution layers (3, 5) and L2 regularization weight (0.0, 0.025) based on the validation performance.

\begin{figure}[t]
\centering
\def\factor{0.85}
\hspace{0.8cm}\textbf{RECOLA}
\includegraphics[width=\factor\linewidth]{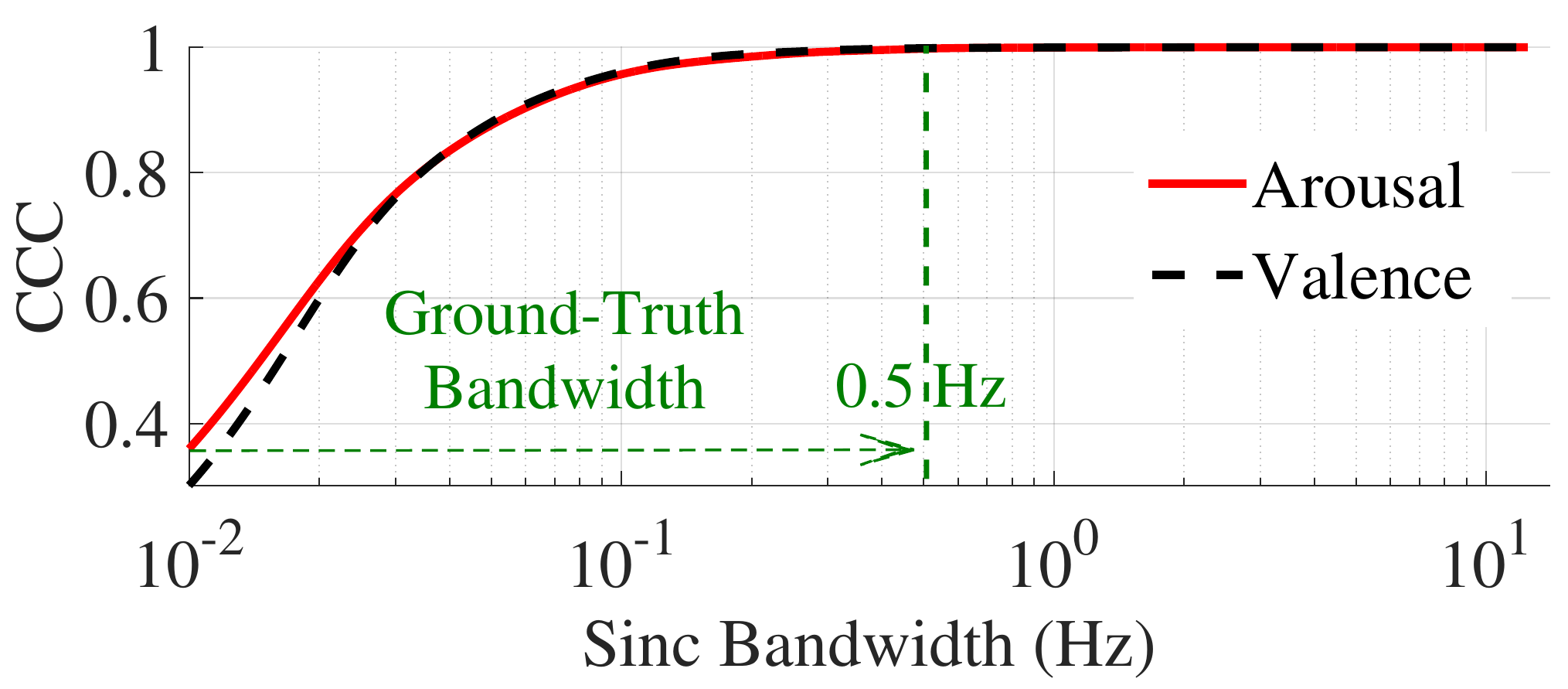}
\\\vspace{0.1cm}\hspace{0.9cm}\textbf{SEWA}
\includegraphics[width=\factor\linewidth]{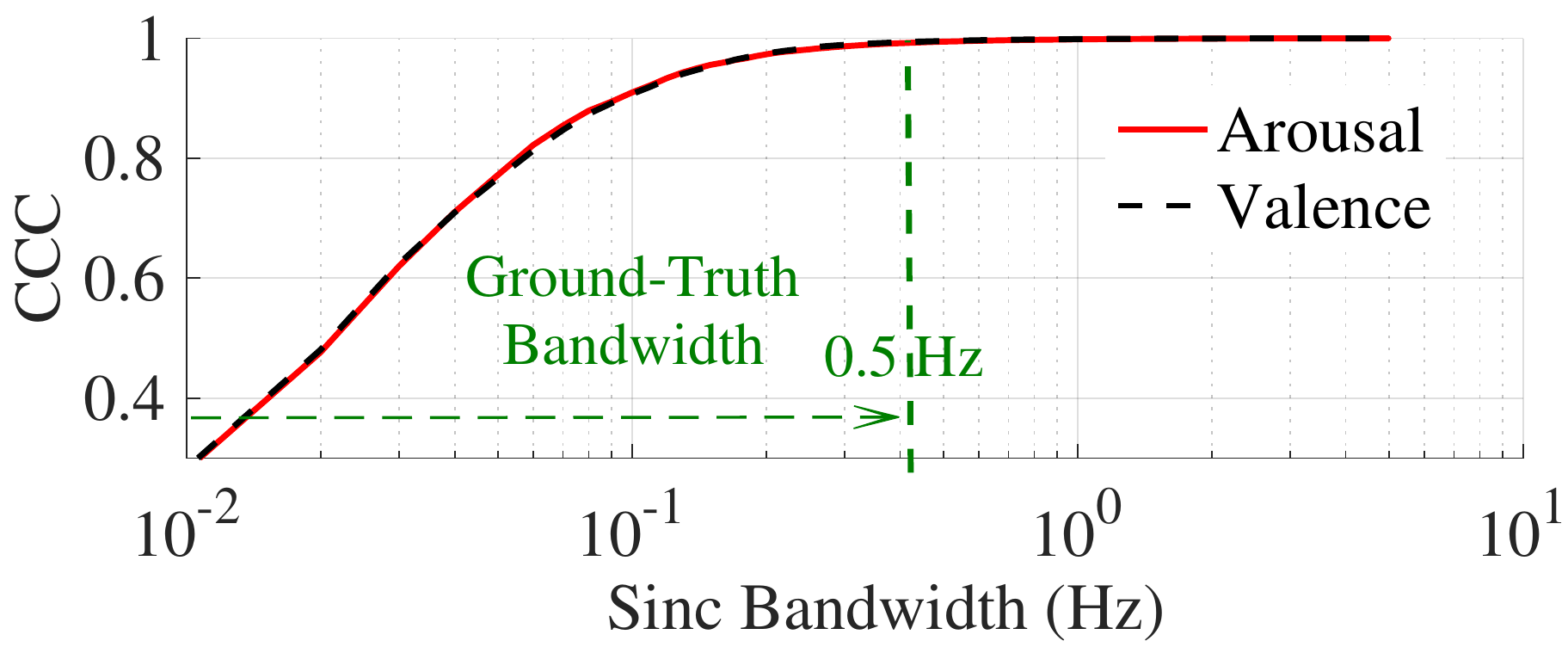}
\caption{Performance reduction caused by applying a windowed-sinc filter to the ground-truth labels on both RECOLA and SEWA datasets. \emph{0.5Hz} can be considered as the bandwidth of the emotion labels.}
\label{fig:effect_of_sinc}
\end{figure}

We conduct an experiment to select a good value for $f_{c}$, the cutoff frequency of the sinc kernels. We apply a windowed-sinc filter with different $f_{c}$ to all training labels and compare the resulting labels with the original ones using the CCC metric (Figure~\ref{fig:effect_of_sinc}). The results obtained for RECOLA and SEWA are very similar, showing that the ground-truth labels in both datasets share similar frequency characteristics; for example, selecting $f_{c}$ greater than 0.5Hz results in a less than 1 percent reduction in CCC on both datasets. Therefore, a number higher than 0.5Hz is a good choice for $f_{c}$. We set $f_{c}$ to $f_{s}/2$ in our experiments, where $f_{s}$ is the sampling rate of the emotion labels (25Hz for RECOLA and 10Hz for SEWA).

We apply 32 delayed sinc kernels of length of 44 seconds. We discuss the effects of kernel length on performance in Section~\ref{sec: Structural Analysis of MDS Network}. We initialize the delay parameters $\tau_m$ of the sinc kernels randomly, between 0 and 20 seconds. Our initial experiments show that the uniform distribution is better that the Gaussian distribution for initializing the delay values. We find that when we initialize the delay values with negative numbers, they tend to converge to positive numbers.  This supports the claim that the delay values approximate the reaction delays of annotators.


\subsection{Results}\label{sec: Results}



We present the results obtained for the RECOLA and the SEWA datasets separately in this section.

\subsubsection{RECOLA}
Table~\ref{tab:leave-one-speaker-out} (RECOLA) reports the leave-one-speaker-out CCCs calculated for both MDS and downsampling/upsampling networks. The results show that our system is significantly better than the downsampling/upsampling network for both arousal and valence. Our arousal predictions exhibit 0.02$\pm$0.024 improvement (pairwise t-test, p-value=$0.04$). Our valence predictions show the improvement of 0.056$\pm$0.07 (pairwise t-test, p-value=$0.047$).


We also report the test CCC results calculated according to the AVEC evaluation scheme explained in section~\ref{Evaluation}. We rank all trained systems in the previous experiment based on their development CCC. We then select the best performing MDS and downsampling/upsampling networks based on the development CCC. We use the selected systems to generate the emotion labels for test utterances. For both the arousal and valence prediction tasks, the best MDS network contains 5 layers with 16 filters in each layer and 8 filter coefficients for each filter. The best downsampling/upsampling network for arousal has 7 layers, 32 filters in each layer and 3 filter coefficients for each filter. It also uses L2 regularization factor of 0.02. The best for valence has 7 layers, 128 filters, 3 filter coefficients and 0.04 L2 regularization factor. 

Table~\ref{tab:test_avec} (RECOLA) summarizes the development and test results calculated according to the AVEC evaluation scheme. For the arousal prediction, all results are slightly in favor of the proposed system. For the valence prediction, our proposed system improves CCC values but it cannot improve RMSE values. We hypothesize that it is because we train the networks to maximize the CCC metric, which does not necessarily lead to a good RMSE estimator.

\definecolor{mygray}{gray}{0.9}
\begin{table}
\centering
\caption{CCC results of downsampling/upsampling (down/up) and MDS networks for each subject in RECOLA and SEWA datasets. leave-one-speaker-out evaluation scheme is used to calculate the CCC results for each subject. MDS network fails to improve the downsampling/upsampling network for the shaded subjects. \vspace{-0.2cm}}
\label{tab:leave-one-speaker-out}
\vspace{0.2cm}\hspace{0.9cm}{\fontsize{9}{9}\selectfont{RECOLA}}\vspace{0.15cm}
\begin{tabular}{c|cc|cc}
\multicolumn{1}{c}{ } & \multicolumn{2}{c}{Arousal}               & \multicolumn{2}{c}{Valence}                  \\
Sub                   & Down/up           & MDS-net               & Down/up               & MDS-net              \\ \hline
1                     & .812              & \textbf{.820}         & .429                  & \textbf{.494}        \\ 
2                     & .922              & \textbf{.938}         & .365                  & \textbf{.440}        \\
3                     & .800              & \textbf{.835}         & .370                  & \textbf{.493}        \\
4                     & .755              & \textbf{.787}         & .135                  & \textbf{.163}        \\ 
5                     & .812              & \textbf{.875}         & .399                  & \textbf{.485}        \\
6                     & .821              & \textbf{.858}         & .445                  & \textbf{.463}        \\
7                     & \textbf{.846}     & .844                  & .265                  & \textbf{.447}        \\
\cellcolor{mygray} 8  & \cellcolor{mygray} \textbf{.637}          & \cellcolor{mygray} .615                      & \cellcolor{mygray} \textbf{.706}                               & \cellcolor{mygray} .657 \\
9                     & .744              & \textbf{.762}         & \textbf{.658}         & .634                 \\ \hline
$\mu\pm\sigma$        & .794$\pm$.08      & \textbf{.814$\pm$.09} & .419$\pm$.18          & \textbf{.475$\pm$.14}\\
\end{tabular}
\\\vspace{0.5cm}
\hspace{0.9cm}{\fontsize{9}{9}\selectfont{SEWA}}\vspace{0.15cm}
\begin{tabular}{c|cc|cc}
\multicolumn{1}{c}{ } & \multicolumn{2}{c}{Arousal}                  & \multicolumn{2}{c}{Valence}             \\
Sub                   & Down/up           & MDS-net                  & Down/up           & MDS-net             \\ \hline
1                     & .368              & \textbf{.488}            & .471              & \textbf{.671}       \\ 
2                     & .325              & \textbf{.400}            & \textbf{.228}     & .218                \\
\cellcolor{mygray} 3                     & \cellcolor{mygray} \textbf{.174}     & \cellcolor{mygray} -.036                    & \cellcolor{mygray} \textbf{.184}     & \cellcolor{mygray} .043                \\
4                     & .372              & \textbf{.579}            & .455              & \textbf{.697}       \\ 
\cellcolor{mygray} 5                     & \cellcolor{mygray} \textbf{.512}     & \cellcolor{mygray} .510                     & \cellcolor{mygray} \textbf{.480}     & \cellcolor{mygray} .454                \\
6                     & .336              & \textbf{.646}            & .373              & \textbf{.662}       \\
7                     & .616              & \textbf{.682}            & .532              & \textbf{.601}       \\
8                     & \textbf{.566}     & .529                     & .350              & \textbf{.601}       \\
9                     & .031              & \textbf{.033}            & \textbf{.062}     & .037                \\
10                    & .474              & \textbf{.525}            & .286              & \textbf{.472}       \\
11                    & .297              & \textbf{.373}            & .321              & \textbf{.501}       \\
\cellcolor{mygray} 12                    & \cellcolor{mygray} \textbf{.241}     & \cellcolor{mygray} .153                     & \cellcolor{mygray} \textbf{.396}     & \cellcolor{mygray} .351                \\
13                    & .516              & \textbf{.615}            & .170              & \textbf{.327}       \\
14                    & .412              & \textbf{.567}            & .392              & \textbf{.406}       \\ \hline
$\mu\pm\sigma$        & .374$\pm$.16	  & \textbf{.433$\pm$.23} & .336$\pm$.14	 & \textbf{.432$\pm$.22}  \\
\end{tabular}

\end{table}

\subsubsection{SEWA}

Table~\ref{tab:leave-one-speaker-out} (SEWA) compares the CCC result of MDS with downsampling/upsampling networks for each subject. Similar to the RECOLA dataset, our system outperforms the downsampling/upsampling network for most of the subjects (10/14 subjects for predicting arousal and 9/14 subjects for predicting valence). Our system exhibits an improvement of $0.059\pm0.13$ on arousal prediction and $0.096\pm0.13$ on valence recognition. The improvement is not significant for the arousal prediction (p-value=$0.11$), but it is significant for the valence prediction (p-value=$0.02$).

\setlength{\dashlinedash}{1pt}
\setlength{\dashlinegap}{1pt}
\setlength{\arrayrulewidth}{0.5pt}
\begin{table}
\centering
\caption{Comparing down/up and MDS networks on RECOLA and SEWA datasets. AVEC evaluation scheme is used for this comparison. \vspace{-0.2cm}}
\label{tab:test_avec}
\def\arraystretch{1.2}\tabcolsep=0.1cm
\vspace{0.1cm}
\begin{tabular}{cc|cc|cc}
\multicolumn{2}{c}{\multirow{2}{*}{\fontsize{9}{9}\selectfont{RECOLA}}}     & \multicolumn{2}{c}{Development} & \multicolumn{2}{c}{Test}       \\
\multicolumn{2}{c|}{}                    & Down/up       & MDS-net                         & Down/up       & MDS-net        \\ \hline
\multirow{2}{*}{CCC}    & Arousal        & .865          & \textbf{.873}                   & .680          & \textbf{.688}  \\
                        & Valence        & .574          & \textbf{.591}                   & .472          & \textbf{.492}  \\ \hdashline
\multirow{2}{*}{RMSE}   & Arousal        & .098          & \textbf{.097}                   & .141          & \textbf{.136}  \\
                        & Valence        & \textbf{.105} & .119                            & \textbf{.116} & .126           \\
\end{tabular}
\\\vspace{0.4cm}
\begin{tabular}{cc|cc|cc}
\multicolumn{2}{c}{\multirow{2}{*}{\fontsize{9}{9}\selectfont{SEWA}}}                  & \multicolumn{2}{c}{Development}  & \multicolumn{2}{c}{Test}  \\
\multicolumn{2}{c|}{}              & Down/up      & MDS-net           & Down/up  & MDS-net        \\\hline
\multirow{2}{*}{CCC}    & Arousal                  & .458         & \textbf{.530}     & .317     & \textbf{.412} \\
                        & Valence                  & .485         & \textbf{.542}     & .287     & \textbf{.379} \\\hdashline
\multirow{2}{*}{RMSE}   & Arousal                  & .139         & \textbf{.135}     & .130     & \textbf{.124} \\
                        & Valence                  & .157         & \textbf{.138}     & .178     & \textbf{.130} \\
\end{tabular}
\end{table}

\setlength{\dashlinedash}{1pt}
\setlength{\dashlinegap}{1pt}
\setlength{\arrayrulewidth}{0.5pt}
\begin{table}
\centering
\caption{Comparing emotion recognition systems on the test set of SEWA.}
\label{tab:test_ccc_sewa}
\def\arraystretch{1.2}\tabcolsep=0.2cm
\begin{tabular}{c|cc|cc}
\multicolumn{1}{c}{  }                       & \multicolumn{2}{c}{Arousal}   & \multicolumn{2}{c}{Valence}\\
Methods                                      & RMSE          & CCC           & RMSE          & CCC            \vspace{0.1cm}\\
\hline
Down/up                                      & .130          & .317            & .178          & .287\\
eGeMAPS-GMR~\cite{dang2017investigating}     & --            & .344            & --            & .346\\
MDS-net                                      & .124          & .412            & .130          & .379\\
IS10-LSTM~\cite{chen2017multimodal} & \textbf{.100} & .422            & \textbf{.112} & .405\\
IS10-LSTM + MDS-net                               & \textbf{.101}          & \textbf{.458}   & \textbf{.114}          & \textbf{.421}
\end{tabular}
\end{table}

\setlength{\dashlinedash}{0.5pt}
\setlength{\dashlinegap}{0.5pt}
\setlength{\arrayrulewidth}{0.5pt}
\begin{table}
\centering
\caption{Comparing space complexity of the best networks trained on both SEWA and RECOLA. Two numbers are reported in each cell: number of training parameters and applied L2 regularization factor. For example, 56K(.02) shows the best network has 56,000 parameters and is trained with an L2 factor of 0.02.}
\label{tab:complexity}
\def\arraystretch{1.2}\tabcolsep=0.2cm
\begin{tabular}{c|cc|cc}
\multicolumn{1}{c}{ }       & \multicolumn{2}{c}{RECOLA}   & \multicolumn{2}{c}{SEWA}\\
Networks    & Arousal             & Valence             & Arousal                & Valence  \vspace{0.0cm}\\
\hline
Down/up     & 56K (.02)          & 703K (.04)         & 290K (.04)            & 1,415K (.04)\\
MDS-net     & \textbf{37K (0)}   & \textbf{37K (0)}   & \textbf{119K (.025)}  & \textbf{57K (.025)} \\
\end{tabular}
\vspace{-0.4cm}
\end{table}

\begin{figure*}[t]
\centering
\def\factor{0.9}
\includegraphics[width=\factor\linewidth]{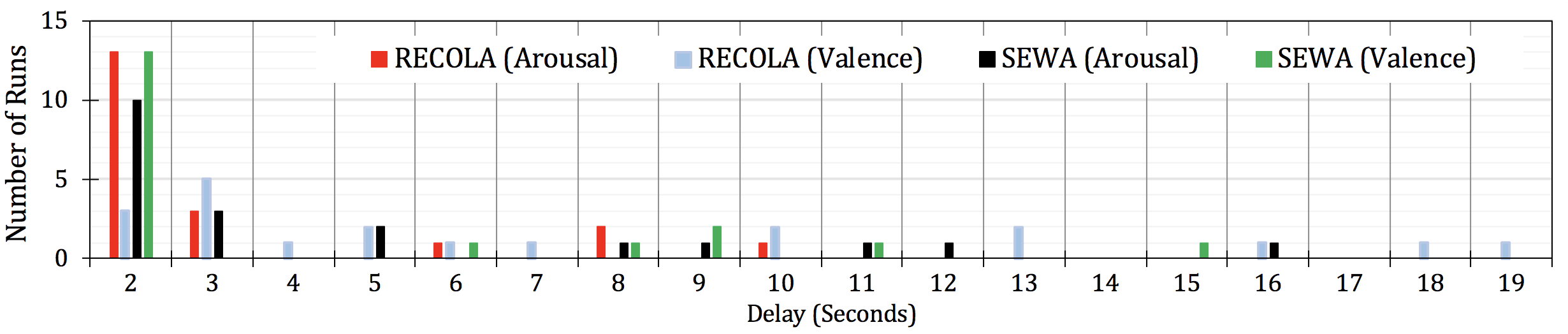}
\vspace{-0.3cm}
\caption{Distribution of the delays trained through different runs of the MDS network with one cluster. This network learns one delay to compensate for the reaction lags.\vspace{-0.3cm}}
\label{fig:one_delay_distribution}
\end{figure*}


We also assess the performance of MDS and downsampling/upsampling networks using AVEC evaluation scheme, explained in section~\ref{Evaluation}. We train our networks with different hyper-parameters explained in section~\ref{Experimental Setup} on the training partition of SEWA. We then select the best performing network based on the development CCC. 

The best MDS arousal predictor contains 3 layers with 32 filters of length 4. This MDS network is trained with L2 regularization factor of 0.025. The best MDS valence predictor differs in only one parameter: it has 16 filters in each layer. The best downsampling/upsampling network for arousal prediction contains 64 convolutional filters of length 3 in each layer with L2 factor of 0.04. For valence prediction, it has 128 convolutional filters of length 5 with L2 factor of 0.04.

Table~\ref{tab:test_avec} shows the development and test results of the selected networks. This table confirms that our network outperforms the downsampling/upsampling network for both arousal and valence prediction tasks on both development and test partitions of the database using both comparison metrics (i.e., RMSE and CCC).

However, the downsampling/upsampling network is not a state-of-the-art method on the SEWA dataset. We also compare the performance of the proposed MDS network with two other emotion recognition systems: eGeMAPS-GMR~\cite{dang2017investigating} and IS10-LSTM~\cite{chen2017multimodal}. Dang et al.~\cite{dang2017investigating} used eGeMAPS features~\cite{eyben2016geneva} with Gaussian mixture regression (GMR)~\cite{metallinou2011tracking} to recognize emotion labels in SEWA. Chen et al.~\cite{chen2017multimodal} used the IS10~\cite{schuller2011recognising} feature set to train a multi-task LSTM network that predicts both arousal and valence labels simultaneously. We note that both systems outperform the downsampling/upsampling network and that while MDS outperforms the eGeMAPS-GMR system, it is outperformed by the IS10-LSTM system (Table~\ref{tab:test_ccc_sewa}).


We analyzed the predictions generated by MDS and IS10-LSTM networks and observed that although both networks are accurate, the generated predictions are often not highly correlated\footnote{Many thanks to the authors of the IS10-LSTM~\cite{chen2017multimodal} paper for sending us predictions of their systems.} (Their CCC for arousal and valence are $0.282$ and $0.362$, respectively; this correlation is calculated on the development set). This suggests that the two approaches may be considering different aspects of the input signal when predicting the labels. We hypothesized that we could improve the predictions by fusing the results of the two systems. We performed the fusion by taking the average of the predictions. We refer to this system as the ``IS10-LSTM + MDS-net" system and find that this system considerably improves the CCC of both ``IS10-LSTM'' and ``MDS-net'' systems while preserving the RMSE of the ``IS10-LSTM'' network (Table~\ref{tab:test_ccc_sewa}). 

Table~\ref{tab:complexity} compares the memory complexity of our best downsampling/upsampling and MDS networks. The number of parameters and the L2 regularization factor of each network are shown in the table. According to the table, the downsampling/upsampling structure requires a larger network with higher regularization factor to predict emotion labels of both RECOLA and SEWA. This high regularization factor is crucial for the large downsampling/upsampling networks to reduce their over-fitting problem. The table confirms that the MDS network outperforms the downsampling/upsampling network with fewer parameters.

\subsection{Structural Analysis of MDS Network}\label{sec: Structural Analysis of MDS Network}
In this section, we analyze various aspects of the MDS network. We try to answer the following questions:
\begin{itemize}[leftmargin=*]
    \item Is delayed sinc layer able to compensate for reaction lags and synchronize speech with emotion labels? (Sec.~\ref{sec: learning delay through delayed sinc layer})
    \item What is the effective range for the bandwidth parameter in delayed sinc layers? (Sec.~\ref{sec: sinc bandwidth})
    \item How many clusters are required to train a robust emotion recognition system? (Sec.~\ref{sec: number of clusters})
    \item What is the maximum delay component that can affect continuous emotion recognition? (Sec.~\ref{sec: maximum delay parameter})
    \item Do the reaction delays change with different acoustic events? (Sec.\ref{sec: Effect of acoustic events on delays})
\end{itemize}

\subsubsection{Learning delay through delayed sinc layer}\label{sec: learning delay through delayed sinc layer}

In Section~\ref{Preliminary Experiment}, we found an estimate of the reaction delay using a brute-force algorithm.
The algorithm trains a separate network for any candidate value of delay and selects the delay that leads to the best CCC result.
This approach is resource intensive and also it is not suitable for training multiple delays.
We demonstrate that the delayed sinc layer can learn comparable delays in a less resource intensive manner using the back-propagation algorithm.

We train our MDS network to learn a single delay by fixing the number of clusters to one (one delay parameter).  We will compare the learned delay to the delay found in Section~\ref{Preliminary Experiment}. We train twenty networks with random initializations and study the distribution of the learned delay values.
We expect that the final delay values will be similar to the values we obtained in Section~\ref{Preliminary Experiment}.
We set all parameters, except the number of clusters, of the networks (e.g., number of layers, length of filters, etc.) to the parameters of the best networks introduced in Section~\ref{sec: Results}.

Figure~\ref{fig:one_delay_distribution} shows the distribution of the trained delays for both arousal and valence predictions on both RECOLA and SEWA datasets.  Although we initialize the delays randomly between 0 to 20 seconds, most of the delays tend to converge to a number between $1.5$ and $3.5$ seconds.  This interval agrees with the optimal delays obtained through exhaustive search in Section~\ref{Preliminary Experiment}.  

It is important to quantify the likelihood of finding an effective reaction delay (a delay between $1.5$ and $3.5$ seconds) using delayed sinc kernel. This likelihood measures the ability of the proposed delayed sinc kernel in finding and compensating for the reaction delays. According to Figure~\ref{fig:one_delay_distribution} this likelihood is different for different datasets (RECOLA, SEWA) and different tasks (arousal, valence). We quantify the likelihood by calculating the percentage of runs in which the trained delay parameter is a number between $1.5$ and $3.5$ seconds. This percentage is equal to $80\%$, $45\%$, $65\%$ and $65\%$ for RECOLA-arousal, RECOLA-valence, SEWA-arousal and SEWA-valence, respectively.

The results of this section confirm that the delayed sinc layer is able to learn and compensate the reaction delays in most runs.  However, in some runs the sinc layer does not converge to the optimal solution of the exhaustive search.  It is because our optimization function (CCC) has multiple local optima and gradient-based optimization algorithms may get stuck in the local optima.


\subsubsection{Sinc bandwidth}\label{sec: sinc bandwidth}
\begin{figure}[t]
\centering
\def\factor{0.7}
\textbf{RECOLA}\\
\hspace{-0.9cm}\includegraphics[width=\factor\linewidth]{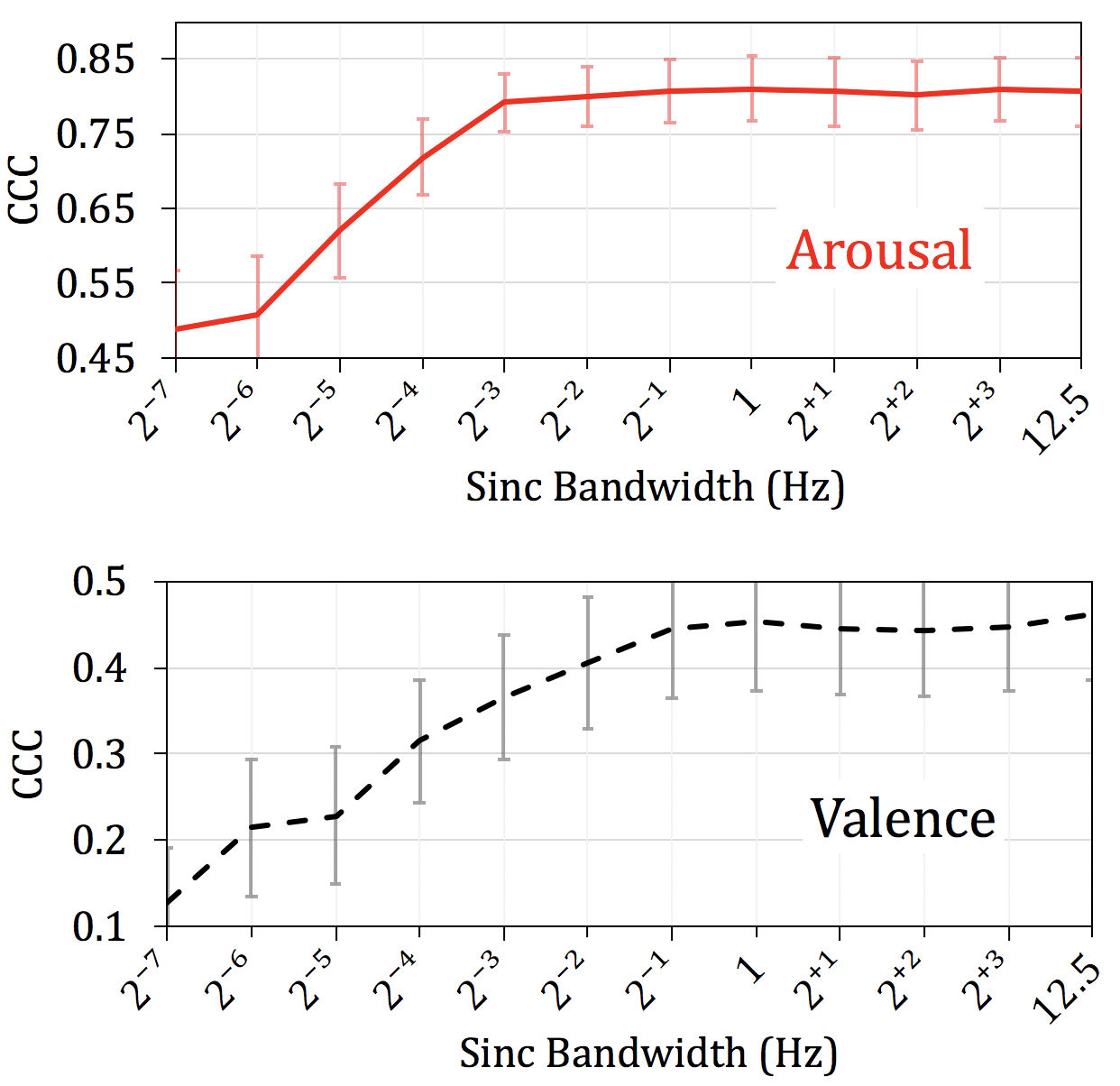}
\vspace{0.2cm}\\\textbf{SEWA}\\
\hspace{-0.9cm}\includegraphics[width=\factor\linewidth]{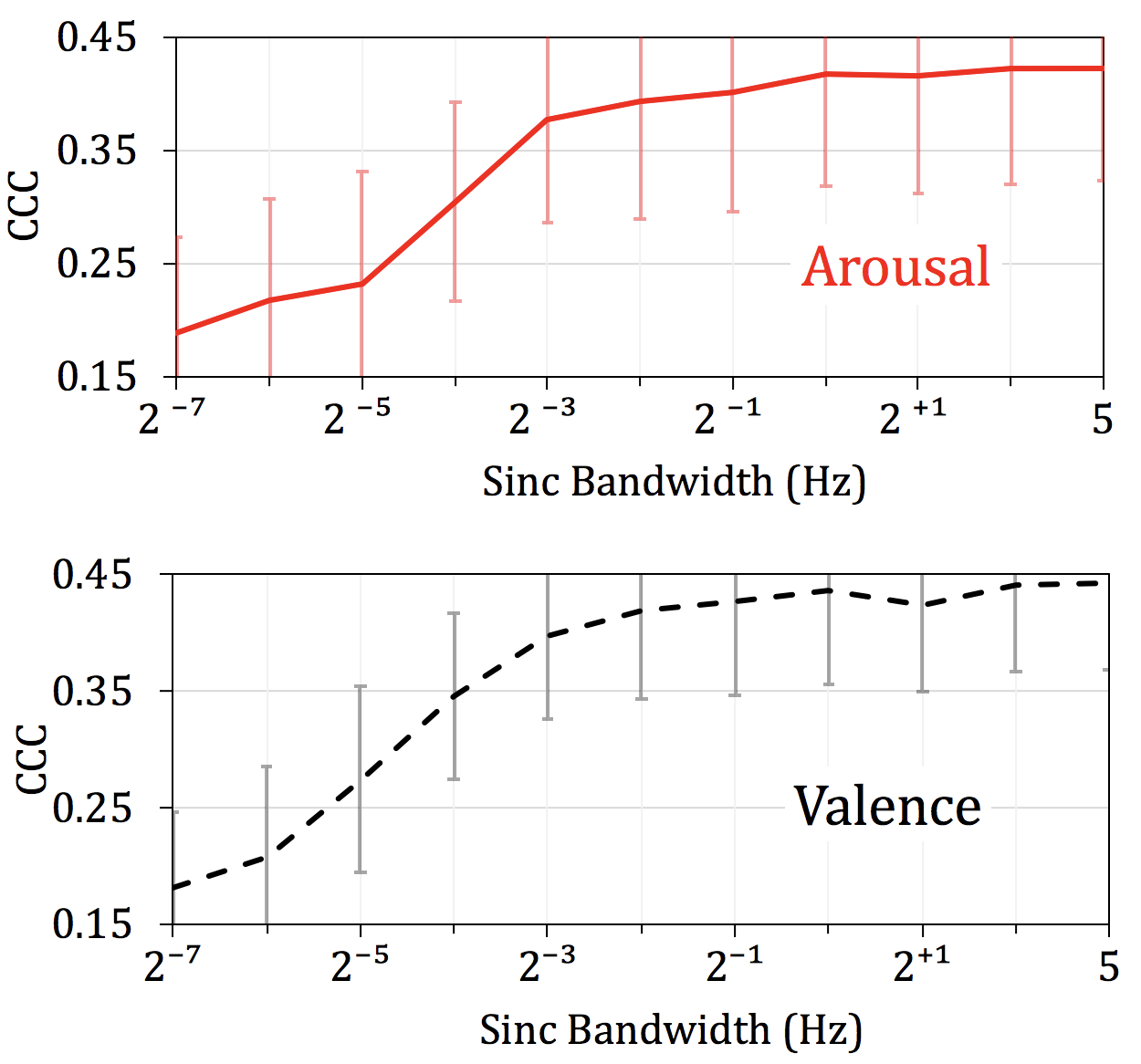}
\caption{Increasing bandwidth of the sinc kernels up to 0.125Hz for arousal and 0.5 Hz for valence improves the leave-one-speaker-out performance. Increasing more does not significantly change the CCC.}
\label{fig:bandwidth}
\end{figure}

The sinc bandwidth parameter can be learned during the training process, but we set it to a constant value (1Hz) throughout the experiments, because our initial experiments showed that learning this parameter does not improve the final CCC. We now explore the relationship between the final performance and the value of this parameter. To this end, we train our best MDS network explained in the previous section with different sinc bandwidths ranging exponentially from $2^{-7}$ to the maximum frequency component (12.5Hz for RECOLA and 5Hz for SEWA). We train each network 10 times with different random initializations and report the average of the leave-one-speaker-out values in Figure~\ref{fig:bandwidth}. As can be seen in the figure, any bandwidth higher than 0.125Hz for arousal and 0.5Hz for valence results in good performance. According to Section~\ref{Delayed sinc Layer} and Figure~\ref{fig:effect_of_sinc}, 0.5Hz is the bandwidth of the continuous emotion labels. Therefore, we do not need to learn the sinc bandwidths during the training process and we can just select a frequency higher than the output bandwidth (i.e., 0.5Hz). Also, Figure~\ref{fig:bandwidth} shows that Valence prediction is more sensitive to attenuating frequencies between 0.125Hz and 0.5Hz.

\subsubsection{Number of the clusters}\label{sec: number of clusters}
\begin{figure}[t]
\centering
\def\factor{0.7}
\textbf{RECOLA}\\
\hspace{-0.75cm}\includegraphics[width=\factor\linewidth]{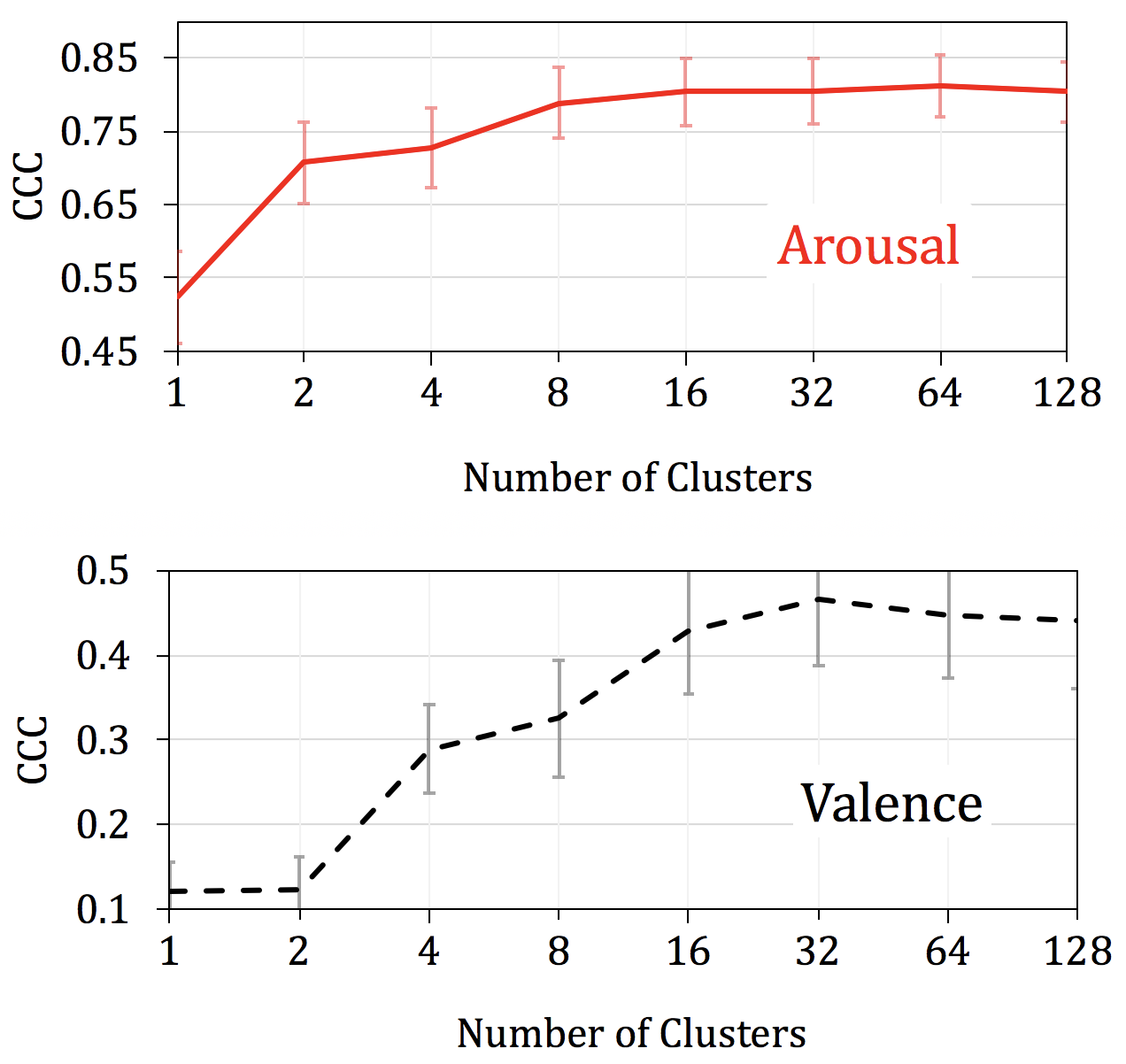}
\vspace{0.2cm}\\\textbf{SEWA}\\
\hspace{-0.75cm}\includegraphics[width=\factor\linewidth]{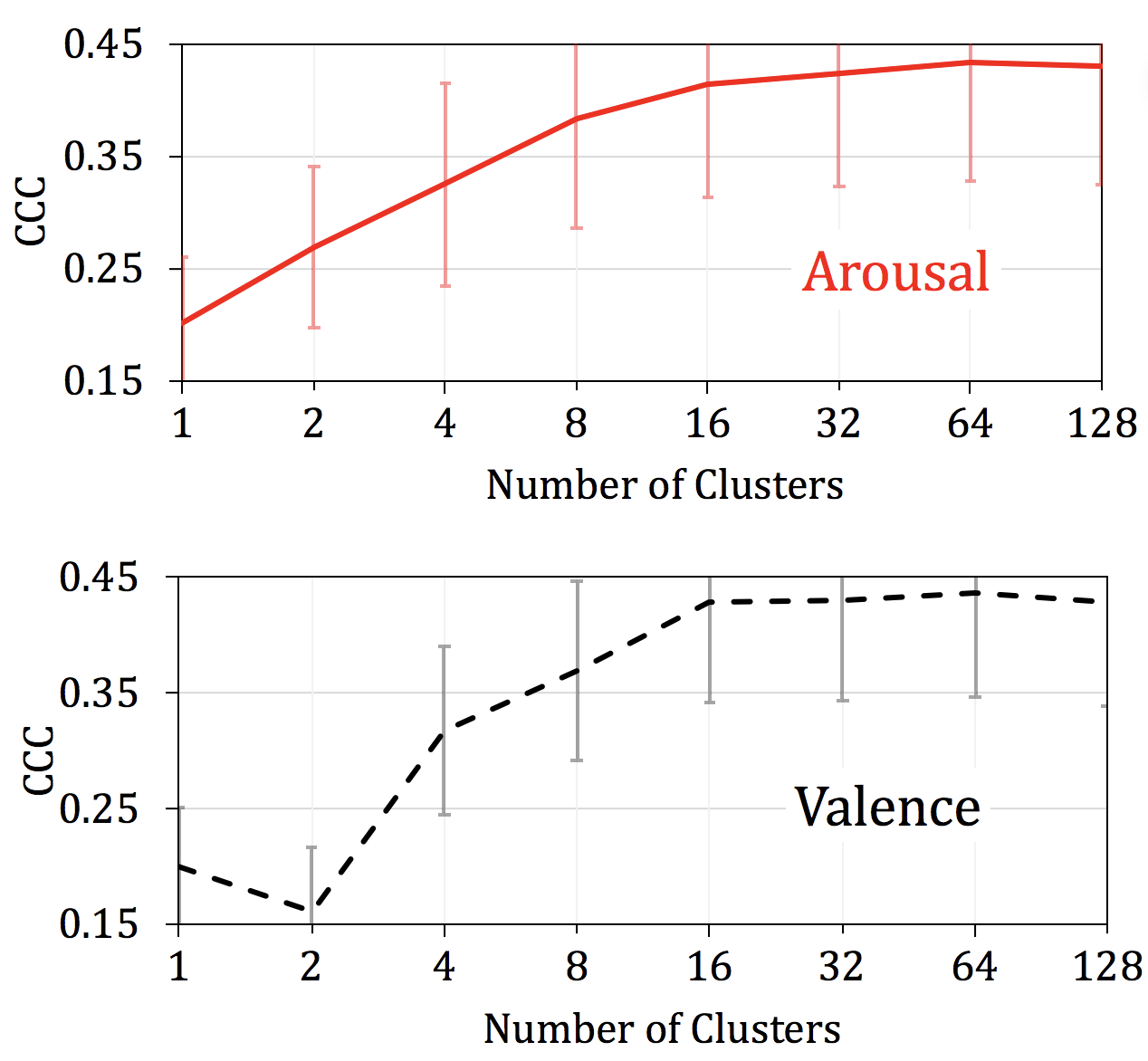}
\caption{CCC results for different number of clusters.\vspace{-0.2cm}}
\label{fig:number_of_clusters}
\end{figure}

Our system divides the acoustic space into several fuzzy clusters and applies different delays to each cluster.  In this section, we investigate the utility of this approach. We train the best MDS network explained in Section~\ref{sec: Results} for different number of clusters ranging exponentially from 1 to 128 and compare their leave-one-speaker-out CCC. Figure~\ref{fig:number_of_clusters} shows the results obtained in this experiment for both RECOLA and SEWA datasets. According to the results, using one cluster is not enough and the MDS network needs at least 8 clusters for arousal prediction and 16 clusters for valence prediction to provide a high performance system that is comparable to the best performing system. The results also show that more clusters are needed to predict valence, compared to arousal.

\subsubsection{Maximum delay parameter}\label{sec: maximum delay parameter}

In this section, our goal is to find the maximum delay that can assist continuous emotion recognition. 
To this end, we train different networks with different maximum delays, $\tau_{max}$, and report the development CCCs of the networks with respect to $\tau_{max}$.
More specifically, we set the maximum initial delay to $\tau_{max}$, the length of delayed sinc kernels to $2\tau_{max}$, and all other hyper-parameters to the best values reported in Section~\ref{sec: Results}.
We train and evaluate the networks on SEWA using the AVEC evaluation scheme explained in Section~\ref{Evaluation}.
We train each network ten times and report the average CCC to reduce the effect of random initialization.

Figure~\ref{fig:delay_laughter_analysis} (solid black curves) shows the results of this experiment for both arousal and valence prediction.
According to this figure, CCC values improve consistently by increasing the maximum delay up to $7.5$ seconds.  
Therefore $7.5$ seconds can be considered as the maximum delay that can assist emotion recognition on SEWA.

\subsubsection{Effect of acoustic events on delays}\label{sec: Effect of acoustic events on delays}
The proposed MDS network compensates for annotators' delays by clustering acoustic space into several fuzzy clusters and by introducing different delays to different clusters; therefore, the network assumes that the delays depend on acoustic clusters. In this section, we explore this assumption, asking if the annotators' delays change with different regions in the acoustic space. 

We study the reaction delays for laughter regions of signal and compare them with the delays of other parts of the signal. We select laughter because it is highly likely that laughter requires smaller reaction delays compared to speech. Many studies discussed acoustic characteristics of laughter and concluded that there are several distinguishable features in laughter; for example, laughter has longer unvoiced portions than voiced portions~\cite{truong2005automatic, bickley1992acoustic}. These features can facilitate identifying laughter and can reduce the annotators' reaction times.

\begin{figure}[t]
\centering
\def\factor{0.75}
\includegraphics[width=\factor\linewidth]{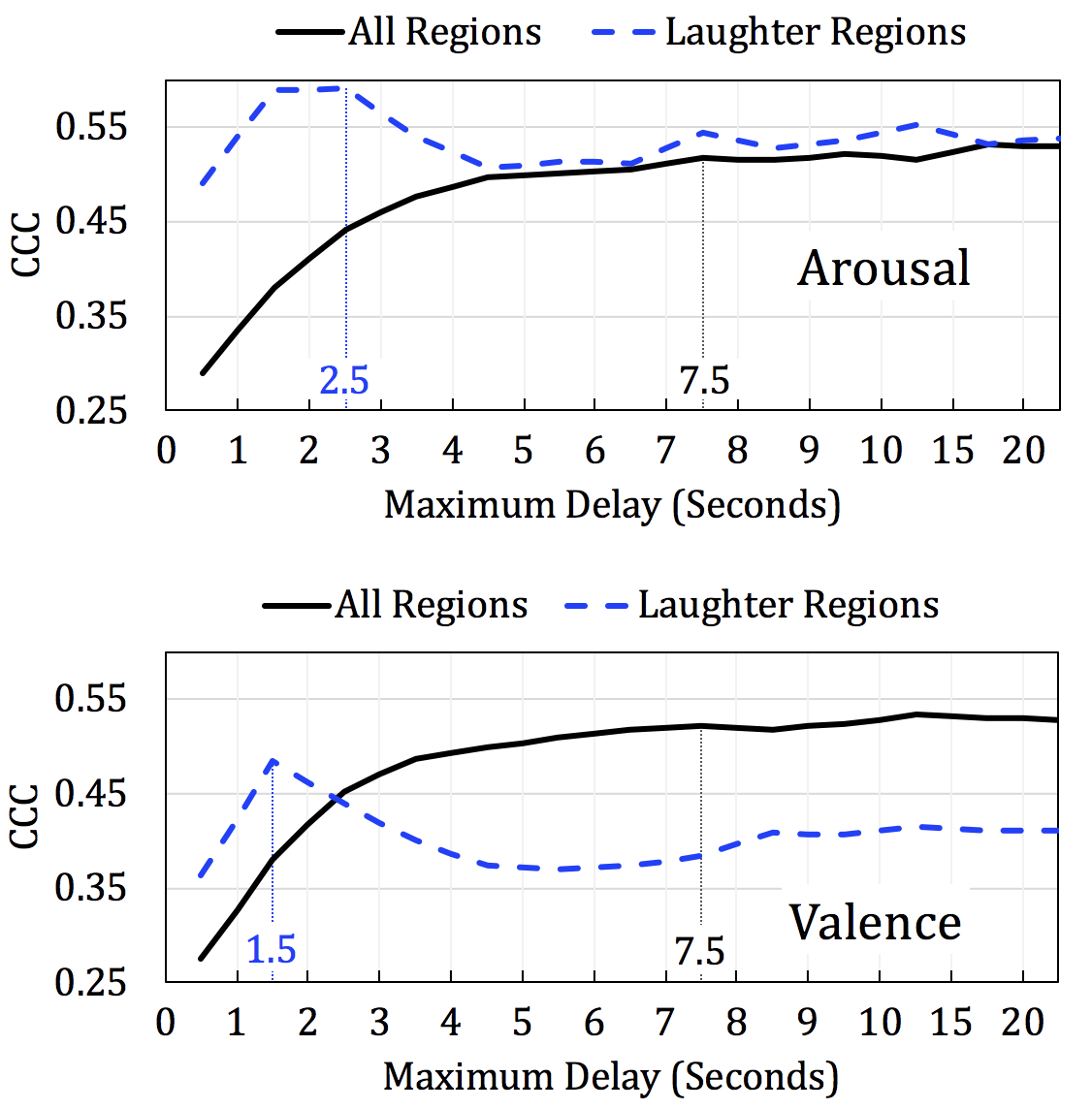}
\caption{CCC results of the proposed network with different maximum delays on the SEWA dataset. AVEC evaluation scheme, explained in Section~\ref{Evaluation}, is used to calculate the CCC values of the laughter parts (dashed blue curves) and all parts (solid black curves) of the development set.}
\label{fig:delay_laughter_analysis}
\end{figure}

In order to find the delays of the laughter parts, we repeat the experiment reported in the previous section, but with the difference that we calculate the CCC results just for laughter parts of the signals. Figure~\ref{fig:delay_laughter_analysis} (dashed blue curves) demonstrates these CCC results with respect to the maximum delay parameter. Increasing the maximum delay parameter more than $2.5$ seconds for arousal and $1.5$ seconds for valence considerably reduces the CCC values. It shows that predicting emotion labels of the laughter regions can be done by applying smaller delay components compared to other parts of a speech signal. We hypothesize that it is because human annotators have smaller reaction delays in identifying emotion labels of the laughter regions and therefore reaction delays depend on acoustic variability.

\section{Discussion}
Annotators' reaction delays make continuous emotion reco\-gnition challenging because the delays introduce non-additive (convolutive) noise components to the emotional labels and create asynchronous ground-truth signals. Compensating for these delays is not straightforward as the delays depend on different factors such as the age of the annotators \cite{der2006age}, concentration level of the annotators \cite{bao2001concentration}, and affective behaviours \cite{mariooryad2015correcting}. 

The proposed network models the reaction delays throu\-gh functions of affective behaviours captured by acoustic clusters. The network categorizes acoustic features into several fuzzy clusters (membership degrees can be any number between zero and one) and applies different delays for different clusters. Since acoustic clusters vary over time, the modeled reaction delay is also time-varying.


The results reported in Table~\ref{tab:leave-one-speaker-out} demonstrate the superiority of our network over the downsampling/upsampling network~\cite{khorram2017capturing}. On the RECOLA dataset, our system is better for most of the subjects (7 out of 9 subjects), but cannot improve the performance on subject 8. We studied the recorded audio files and we noticed that for all files in the development set, the interviewee's voice dominates the leaked interviewer's voice, but for subject 8 the power of the interviewer's voice is comparable to the power of the interviewee's voice. This considerable leakage may be a reason that our system behaves differently for this subject.

The leave-one-speaker-out experiment (Table~\ref{tab:leave-one-speaker-out}) shows that the CCC results of arousal recognition are very different for RECOLA and SEWA datasets (CCC result is around $0.8$ on RECOLA, but around $0.4$ on SEWA).  In addition, the MDS network performs similarly in predicting arousal ($0.433$ CCC) and valence ($0.432$ CCC) on SEWA, but not on RECOLA; predicting arousal is easier on RECOLA. One important reason for these inconsistencies is that RECOLA and SEWA use different methods for collecting emotion labels. The differences between them were discussed in section~\ref{dataset_differences}.

\vspace{-0.2cm}
\section{Conclusion}\label{Conclusion}
This paper introduces a new method to align acoustic features and continuous emotion labels using the delayed sinc layer. This layer is able to introduce a learnable delay to its input. Our experiments show that the delayed sinc layer can successfully align two signals by introducing a single (uniform) delay to one of them. However, a uniform delay is not enough for aligning speech and emotion annotations, because reaction delays of annotators may vary with affective behaviours. To deal with this issue, we combine multiple delayed sinc layers into a network architecture. The network categorizes features into a number of fuzzy clusters (it is fuzzy because each acoustic feature can belong to more than one cluster), and then learns different delays for different acoustic clusters. Our experiments show that: (1) the sinc filter with a cutoff frequency higher than the bandwidth of the ground-truth signal can be used to deal with the misalignment problem; (2) the proposed network significantly outperforms the downsampling/upsampling network; (3) predicting valence requires more clusters compared to predicting arousal. (4) laughter requires smaller delay components compared to other regions of speech. 

We used a sinc filter to approximate the dirac-delta function. However, other functions, such as Gaussian and triangular, can also be employed instead of the sinc kernel. Future work will explore the effect of using different types of kernels that can approximate the dirac-delta function. Additionally, in this paper, we focused on the speech modality to predict continuous emotion annotations, while the proposed multi-delay sinc network is a reasonable modeling technique for other input modalities too. Another future plan is to evaluate the performance of the proposed network over other physiological and behavioral modalities such as: video~\cite{kaya2017video, noroozi2017audio}, body language~\cite{balas2017body, daoudi2017emotion} and EEG~\cite{mert2018emotion, zheng2017identifying}. 

\vspace{-0.2cm}
\section*{Acknowledgment}
This work was partially supported by the National Science Foundation (NSF CAREER 1651740), National Institutes of Health (R34MH100404, R21MH114835, UL1TR002240), HC Prechter Bipolar Program, and the Richard Tam Foundation. 

\vspace{-0.4cm}




\bibliographystyle{IEEEtran}
%

\bibliography{mybib}


\begin{IEEEbiography}[{\includegraphics[width=1in,height=1.25in,clip,keepaspectratio]{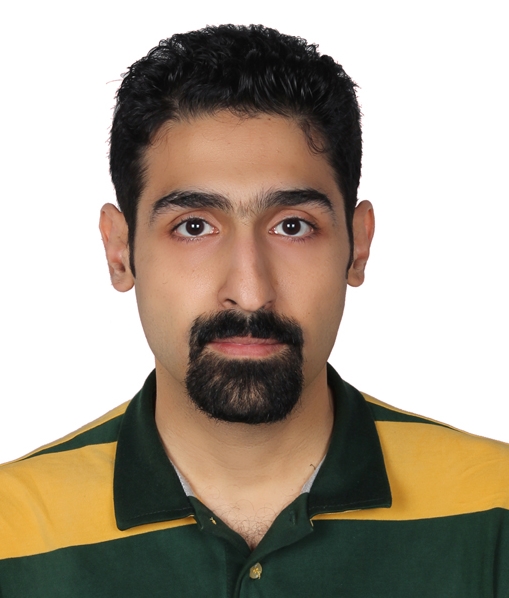}}]%
{Soheil Khorram}
is a Research Fellow in Computer Science and Engineering at the University of Michigan. He received his M.S. and Ph.D. in Computer Engineering from Sharif University of Technology, Tehran, Iran in 2008 and 2015 respectively. His M.S. and Ph.D. thesis was about "speech enhancement using subband adaptive filtering" and "improving acoustic models for statistical parametric speech synthesis". He was also a visiting student at the Center for Speech Technology Research (CSTR), University of Edinburgh, from July to December, 2013. He worked as a member of the Simple4All project under the supervision of Prof. Simon King at CSTR. He is currently working on emotion and mood recognition systems as a part of the PRIORI (Predicting Individual Outcomes for Rapid Interventions) project.\vspace{-1cm}
\end{IEEEbiography}

\begin{IEEEbiography}[{\includegraphics[width=1in,height=1.25in,clip,keepaspectratio]{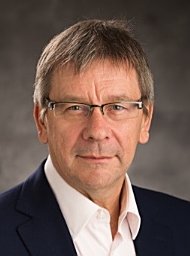}}]%
{Melvin G McInnis} MD, is the Thomas B and Nancy Upjohn Woodworth Professor of Bipolar Disorder and Depression and Professor of Psychiatry. He is the Director of the HC Prechter Bipolar Program and Associate Director of the Depression Center at the University of Michigan. He is a Fellow of the Royal College of Psychiatry (UK) and Fellow of the American College of Neuropsychopharmacology. Dr. McInnis trained in Canada, Iceland, England, and USA, he began a faculty position in Psychiatry at Johns Hopkins University (1993) and was recruited to the University of Michigan in 2004. His research interests include the genetics of bipolar disorder and longitudinal outcome patterns in mood disorders. He has received awards recognizing excellence in bipolar research from the National Alliance for the Mentally Ill (NAMI) and National Alliance for Research in Schizophrenia and Affective Disorders (NARSAD). He has published over 250 manuscripts related to mood disorders research, and is widely engaged in collaborative research focused on identifying biological mechanisms of disease and predictive patterns of outcomes in mental health.

\end{IEEEbiography}


\begin{IEEEbiography}[{\includegraphics[width=1in,height=1.25in,clip,keepaspectratio]{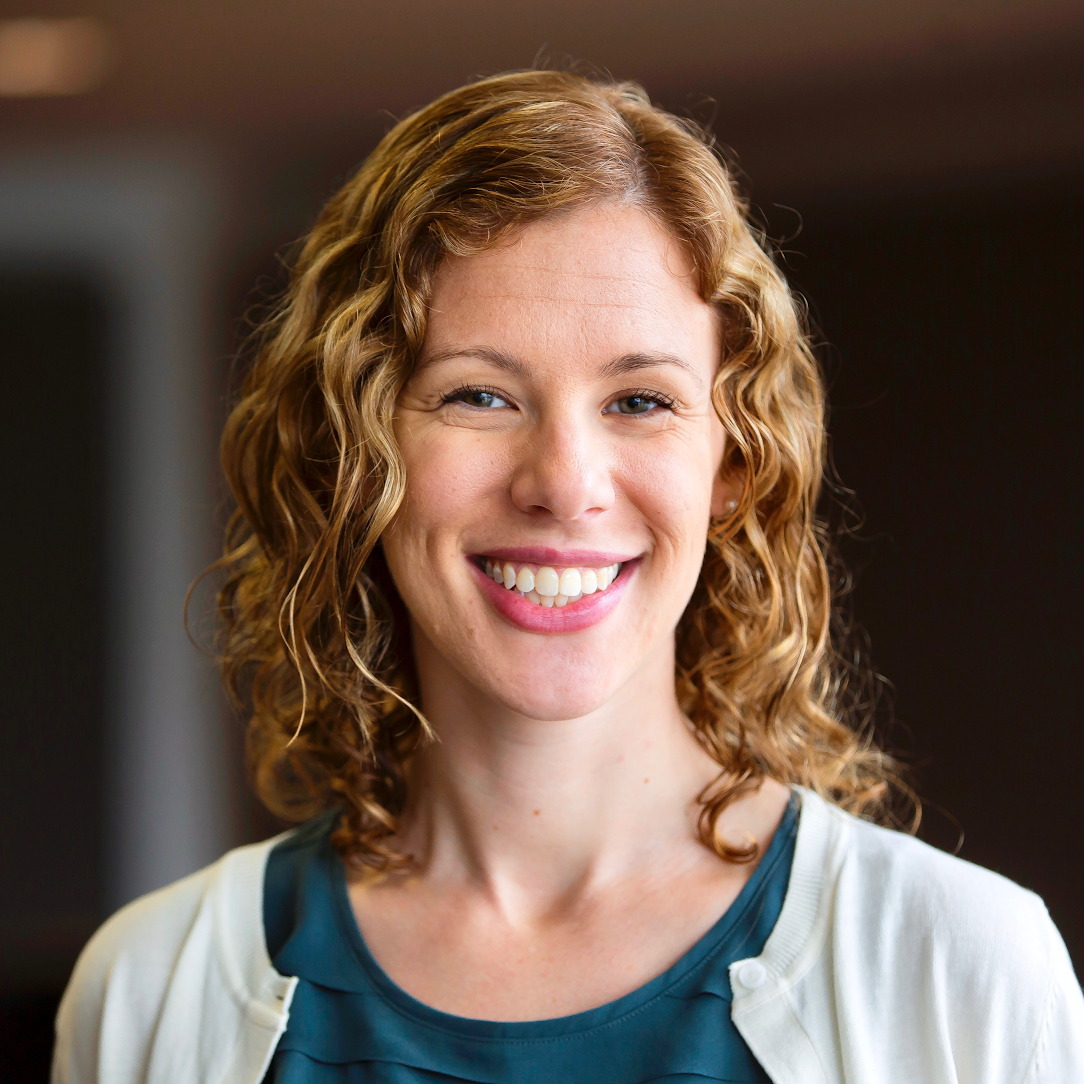}}]%
{Emily Mower Provost}
   is an Associate Professor in Computer Science and Engineering at the University of Michigan. She received her B.S. in Electrical Engineering (summa cum laude and with thesis honors) from Tufts University, Boston, MA in 2004 and her M.S. and Ph.D. in Electrical Engineering from the University of Southern California (USC), Los Angeles, CA in 2007 and 2010, respectively. She is a member of Tau-Beta-Pi, Eta-Kappa-Nu, and a member of IEEE and ISCA. She has been awarded a National Science Foundation CAREER Award (2017), a National Science Foundation Graduate Research Fellowship (2004-2007), the Herbert Kunzel Engineering Fellowship from USC (2007-2008, 2010-2011), the Intel Research Fellowship (2008-2010), the Achievement Rewards For College Scientists (ARCS) Award (2009 – 2010), and the Oscar Stern Award for Depression Research (2015). She is a co-author on the paper, ``Say Cheese vs. Smile: Reducing Speech-Related Variability for Facial Emotion Recognition,'' winner of Best Student Paper at ACM Multimedia, 2014, a co-author of the winner of the Classifier Sub-Challenge event at the Interspeech 2009 emotion challenge, and a co-author of an honorable mention paper at ICMI 2016. Her research interests are in human-centered speech and video processing, multimodal interfaces design, and speech-based assistive technology. The goals of her research are motivated by the complexities of human emotion generation and perception.
\end{IEEEbiography}


\end{document}